\documentclass[final, 5p,times,twocolumn]{elsarticle/elsarticle}

\usepackage{amssymb,amsmath}
\usepackage{algorithm,algpseudocode}
\usepackage{mathtools, cuted}

\journal{(not yet)}

\begin{document}

\begin{frontmatter}



\title{Prediction of Usage Probabilities of Shopping-Mall Corridors Using Heterogeneous Graph Neural Networks}


\author[label1]{Malik M. Barakathullah\corref{cor1}}
\author[label2]{Immanuel Koh\corref{cor2}}
\cortext[cor2]{Email: immanuel\_koh@sutd.edu.sg}
\cortext[cor1]{Email: dr.malik.barak@gmail.com}

\affiliation[label1]{organization={Singapore University of Technology and Design, Architecture and Sustainable Design},
            addressline={8 Somapah Road}, 
            city={Singapore},
            postcode={487372}, 
            country={Singapore}}

\begin{abstract}
We present a method based on graph neural network (GNN) for prediction of probabilities of usage of shopping-mall corridors. The heterogeneous graph network of shops and corridor paths are obtained from floorplans of the malls by creating vector layers for corridors, shops and entrances. These are subsequently assimilated into nodes and edges of graphs. The prediction of the usage probability is based on the shop features, namely, the area and usage categories they fall into, and on the graph connecting these shops, corridor junctions and entrances by corridor paths. Though the presented method is applicable for training on datasets obtained from a field survey or from pedestrian-detecting sensors, the target data of the supervised deep-learning work flow in this work are obtained from a probability method. We also include a context-specific representation learning of latent features. The usage-probability prediction is made on each edge, which is a connection by a section of corridor path between the adjacent nodes representing the shops or corridor points. To create a feature for each edge, the hidden-layer feature vectors acquired in the message-passing GNN layers at the nodes of each edge are averaged and concatenated with the vector obtained by their multiplication. These edge-features are then passed to multilayer perceptrons (MLP) to make the final prediction of usage probability on each edge. The samples of synthetic learning dataset for each shopping mall are obtained by changing the shops' usage and area categories, and by subsequently feeding the graph into the probability model. 

When including different shopping malls in a single dataset, we also propose to consider graph-level features to inform the model with specific identifying features of each mall.
\end{abstract}




\end{frontmatter}


\section{Introduction}
\label{intro}

Due the conveniences such as an integrated parking space, restrooms and thermal comforts a shopping mall provide, the behaviour of general shoppers in cities has been greatly transformed in order to save time and to enhance their shopping experience. However, the footfall into a mall greatly depends on the features of the shops and amenities in it, besides the accessibility of the mall itself through transport networks and the other external amenities in the proximity outside the mall.

In the present work we are interested in a parameter related to quantifying the pedestrian movements that captures a steady-state or time-averaged behaviour on a graph network of one-dimensional paths, i.e. the corridors of a shopping-mall level represented as a string of connected line segments spanning a two dimensional space. We identify this parameter as the average \emph{probability of usage} (PoU) of different sections of the corridor path by a mall entrant. Representing the malls as distance-preserving Euclidean graphs with corridor-following edges is a natural selection for this purpose, since one can make use of various graph-based algorithms, for example, to find the shortest path, or to analyze the centrality measures of the nodes and (graph) edges. 

Given that there are $n$ number of entrants to the mall, the goal of this work is to determine what fraction of them would have used a chosen section of the corridor between two adjacent nodes. Here, each node can represent either a shop or a junction of different sections of a corridor. In what follows, we will use the term \emph{edge} following the jargon of graph theory to represent the corridor section between two adjacent nodes. Since we have two types of nodes, namely shop nodes and non-shop nodes, our graph is \emph{heterogeneous}.

The interest in the above problem arises mainly to estimate two important features pertaining to the monetization and compliance. Firstly, it can help estimating the visibility score for each of these shops. This in turn can help estimating the rents. Secondly, from the usage probabilities one also could monitor the critical sections of the corridor path that may violate compliance rules of the regulating bodies for congestion or occupancy limit, and thus could help in making decisions related to changing the type of shops with appropriate usage density, if necessary.

The driving forces in this dynamics are the shops' attraction strength that depends monotonously on its area and usage density. The usage density of a shop is defined in this work as the average number of persons in the unit area of the shop. 

In general, the PoU of a section of the corridor could also depend on the  proximity of different entrances of the mall to other sources of crowd in the vicinity outside the mall. However, the probability model that we adopt for preparing the training dataset assumes that these entrances have equal priori probabilities. Generalizing this to the scenario where certain entrances are more important than the others is straightforward as we will describe in later sections.

The way how each shops has been connected by the corridors to the other shops and entrances too greatly determine the chances of finding a random pedestrian in a chosen corridor. This graph characteristic is quantifiable in terms of the network centrality measures for each nodes or edges, and is considered in our training-dataset preparation by making use of shortest-path algorithms. Further, this characteristic is naturally incorporated in the graph neural networks (GNN) that we adopt for predictions through the adjacency matrix~\cite{goodrich2015algorithm} of the underlying graph. 

Since our target variable lives on the graph edges, it requires a special attention. Most GNN layers such as GCN~\cite{kipf2016semi}, GraphSAGE~\cite{hamilton2017inductive}, GAT~\cite{velivckovic2017graph} or GIN~\cite{xu2018powerful} output features on each nodes after performing their respective version of message-passing algorithm~\cite{gilmer2017neural, battaglia2018relational}. There are several graph layers,for example, GAT, GINE~\cite{hu2019strategies}, or GMM~\cite{monti2017geometric} that have the ability to consider the edge-level features, but they do not output on edges. (For an overview on the performance of these networks on various existing datasets with respect to different tasks such as link-prediction or regression, please see the benchmark in~\cite{dwivedi2023benchmarking}.)

Therefore, we need a mechanism to translate the hidden layer features on each nodes to that on each edge. There are a variety of methods available in literature in the context of \emph{link prediction}, where the task is a binary classification to determine whether any chosen pair of nodes can form an edge. This is performed in a number of ways. 

In one method, for each pair of nodes, the hidden features from the last layer of the GNN block is concatenated to form a feature for that possible edge connecting the two nodes, which are then subsequently handled by a multi-layer perceptron (MLP) followed by a cross-entropy loss function to determine whether they form an edge. In some works, instead of the operation of concatenation followed by an MLP, a contraction through a dot product is considered.~\cite{trouillon2016complex, kipf2016variational} This is an example for known graph auto-encoders (GAE) for link prediction where the dot-product operation followed by a nonlinear switch serves as the decoder. This edge embedding tasks have also been used in the contexts other than link prediction~\cite{battaglia2016interaction, kipf2018neural, battaglia2018relational} where these are subsequently converted into node embeddings in later layers of the model.

In a method known as SEAL, localized subgraphs around the pairs of nodes are used.~\cite{zhang2018link, li2020distance, zhang2021labeling} A line-graph approach also has been used for this task in a recent work~\cite{cai2021line} which also compare these methods through an application. 

In our problem, we do not predict whether an edge exist between a pair of nodes as done in the above link-prediction problem, but we predict a target variable on each of the already existing edges. For this we use a an idea similar to those used for link prediction. Our choice for this task is a the method closely related to GAE described above, except that we will be using a decoder that is different from the one that uses dot-product operation. 

It should be mentioned that such prediction of a target variable on edges have been come across in a previous work~\cite{wei2019dual} where a line-graph approach has been adopted. In this work, however, the input data also have attributes on the edges. Since these edges serve as nodes in the line-graph approach, normal GNN algorithms can be used in predicting the target variables (on edges). 

However, in our case, the edges of the graph, i.e., the sections of shopping-mall corridors do not have attributes of their own. Therefore, our method is similar to that of GAE. In this work we follow the following steps: After the message-passing layers, the hidden features of nodes at both ends of each edge are averaged and concatenated with that obtained from multiplication which are passed through few fully connected layers to output a single scalar which is compared against the target by a mean-squared error (MSE) loss function. It should be noted that nonlinearity of the input features are not needed in neural networks that uses activation functions (see, for example,~\cite{hastie2009elements} page 394). However the product of the node features at the ends of each edge is not the nonlinearity that could be obtained from the moments of the average of the same features. This is further discussed in the subsection~\ref{poe}.  

We adopt the message-passing methods in our work over the non-message-passing methods such as LINKX~\cite{lim2021large} as the former helps in propagation of information via edges as we go to deeper layers. This helps the predicted variables to depend not only on neighbours, but also on $n-$th nearest neighbours, where $n$ is the depth of the GNN model.

Some of the graphs representing small shopping malls in our dataset have quite a small (graph) diameter, which makes them susceptible to over-smoothing problem~\cite{li2018deeper, zhao2019pairnorm, keriven2022not} that is inherent in the message-passing framework. To overcome this issue, we use residual skip connections between layers that joins their outputs nodes either by an addition or a concatenation~\cite{li2019deepgcns}.

Representation learning harvests some latent features that contain some structural information about the graph~\cite{hamilton2017representation}. This is usually performed to increase the feature vector size on the nodes which in turn helps in increasing the learnable parameters in the model. We only have 2 categorical features for shop nodes, namely the usage and the area categories. When one-hot coded, the feature vector size for these shop nodes is 10. The non-shop nodes, which are bridging nodes on the corridor points, do not have any explicit features. Therefore they need to be obtained through a representation learning method. In this work we skip the traditional methods, namely Node2vec~\cite{grover2016node2vec} and factorization methods. 

Node2vec is an ideal choice when working with a single large graph. Since we consider different malls too in a single dataset, Node2vec would not be an appropriate choice, as it would mean in the analogy of Word2vec that the features learnt from one corpus of a language will be having some relation to that of other languages.

We also skip the method of factorization of adjacency matrix~\cite{ahmed2013distributed}. Since the number of nodes of each graph, and their connections are different, the number of features obtained from decomposition of their respective adjacency matrices would also would be different. This makes them unsuitable to be part of same dataset. Like Node2vec, this method is also suitable for situation where there is a single graph.

For the above reasons, we use a method of context-specific feature engineering for the non-shop nodes to obtain 10 features for each of them. Due to such small number of node features, we use several message passing layers to increase the number of learnable parameters. The engineered features contain some statistics on the number of shortest distance paths from the entrances to each of the shop nodes as described in a later section. 

We also consider a graph level features when the training is made on the dataset that include different graphs. These features are statistics on the number of nodes, degrees, and statistical information on the shortest paths.

The method of generation of target data, i.e., the PoU on the graph edges in the training set follows a probabilistic approach. This approach takes into account of the fact that the attraction strength of the shops would be higher if it is categorized as of high usage, and similarly with respect to area. We consider 5 categories for the shop usage and similar number for the area category. Such synthesis of target data in the absence of real data from sensors or survey is not uncommon in the context urban studies.~\cite{de2017novel} 

In the next section we highlight some related works. In the subsequent section titled \emph{Materials and Methods}, we describe the methods of graph extraction, target data generation using a probability approach, feature engineering for non-shop nodes on the corridors, and finally, the model architecture. The later sections discuss the results and conclusion.

\section{Related Works}
Studying the pedestrian flow has been of interest to architects and urban planners for several decades~\cite{schadschneider2002traffic, thalmann2012crowd} due to its relevance in understanding the crowd movements~\cite{helbing1995social}, its control, urban planning~\cite{lam2000pedestrian, yin2013assessing, liu2014agent, lopez2021modeling}, and in analyzing spatial designs of buildings~\cite{kneidl2012generation}. 
\subsection{Mathematical and Probabilistic approaches}
In theoretical studies this phenomenon has been modelled in multiple ways. For instance, the modelling approaches use the formulations of deterministic continuum mechanics of fluid flows, or the discrete approaches such as the cellular automata and agent-based models (ABM). The ABM's could also make use of probabilistic description of the dynamics similar to the method adopted in the training-dataset generation in this paper. For recent reviews of available methods, see, for example, the report in Ref.~\cite{aghamohammadi2020dynamic, yang2020review}. In most of these works, the pedestrians move in two or three dimensions incorporating complexities governed by various forces that act in different length scales, except in the works presenting traffic-flow models which model the routing of vehicles in a network of one-dimensional lines representing the roads. These approaches also have the ability to capture the details of crowd-density function in the case of continuum approach, or the spatio-temporal evolution of individual agents in the case of ABM's. 

\subsection{Machine Learning Approaches}
Besides the continuum and discrete approaches mentioned above, the research community also uses various methods of machine-learning for the pedestrian behaviour as reviewed by Ridel {\it et al}.~\cite{ridel2018literature}. These methods predict the behaviours learnt from datasets that are generated from sensors and cameras. However most of these works do not use graphs in their models. For example, the recent work by Shi {\it et al.} (2022)~\cite{shi2022deep} is concerned with detecting pedestrian trajectories on the walkways of an urban area using 1D-CNN and LSTM models without using the underlying graph of the walkways in their neural network model, though it has been used in the dataset generation. Due to the presence of time-history in the training data, it becomes possible to predict reasonably well as found by this paper. However, for predicting time-averaged or time-accumulated results such as a path's usage-probability as in this current paper, such time-history that guides the learning would be absent in the dataset, and would warrant other feature-driven approach as we consider in our work.

\subsection{GNN for Mobility Predictions or in Built Environments}
Implementation of graphs in the model would be a direction for the researchers to go, since the exploitation of the connections between nodes by pedestrian paths will greatly enhance the predictive behaviour. For an analogy in the context of images, it is important to note that the incredible advent of deep-learning methods over that of image processing for image-like data in the problems of classification and segmentation is due to the fact that the CNN and pooling layers in these deep models make use of the underlying graph of the pixels inherently. The fact that the convolution and pooling are performed on the patches of the image suggest that they inherently use the concept that the involved pixels in these operations have the relationship of being adjacent to other pixels in the patch. This gives a motivation for using graph based deep learning method for the current problem described above, since the degree of usage of a section of a corridor path, i.e., the \emph{edge} depends strongly on the nodes at its ends and that of its neighbouring edges. 

The usage of graph-neural networks (GNN) in prediction of human mobility has been fruitful recently in the context of road networks that connect different cities, or different localities within same cities~\cite{martin2020graph, rico2021graph, terroso2022nation, saenz2023nation}. There are also several works that uses GNN models that uses RNN layers combined with graph algorithm for predicting pedestrian trajectories using time-history dataset.~\cite{huang2019stgat, monti2021dag, mohamed2020social, haddad2021self}. For a review of these works, see, for example,~\cite{korbmacher2022review}. These works concern about predicting the outputs on the nodes. A recent work~\cite{jin2023mobility}, however, considers the correlation between the graph nodes, which is essentially like a link-prediction problem. However the link prediction method adopted in their work is quite different from the framework of our GAE approach, since they use attention networks to infer the link prediction.

Graph representation of corridor paths have been used in the past for pedestrian analysis~\cite{christakos2006simple}, since it enables representing a spatially connected unstructured information as in the floorplans, though that work do not use GNN. Usage of the graphs for representing floorplans has gained momentum in the past few years due to developments and applications of deep learning methods~\cite{hu2020graph2plan}. An earlier work~\cite{yan2019graph} has adopted similar graph representation for a group of buildings in a classification problem using GNN. 

More recently, Yang \& Huang (2023)~\cite{yang2023representation} represented shopping malls as graphs for a GNN-based classification problem. However, the edges of these graphs did not follow the corridors, and neither the underlying graph was Euclidean, since it concerned with graph level classification rather than predictions at node or edge levels.

\section{Dataset Preparation}
\begin{figure*}[htb!]
	\centering
	\includegraphics[width=0.9\linewidth]{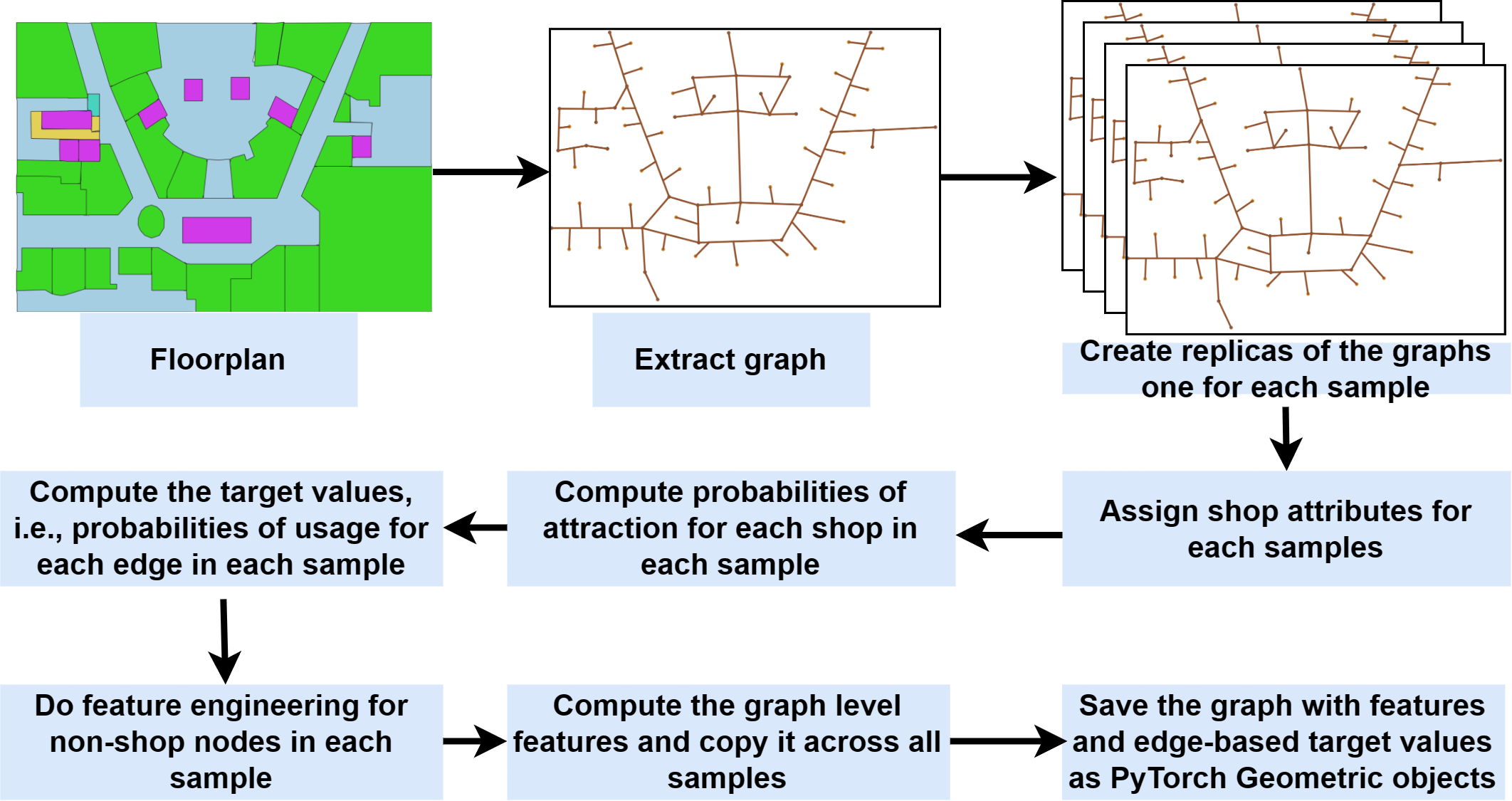}
	\caption{A schematic diagram showing the dataset preparation for a chosen floorplan. This process is repeated for each floorplan and combined into a single set of training and test datasets.}
	\label{fig:schematic}
\end{figure*}

We create a graph database where each mall with a particular arrangement of shops' usage (i.e., the level of service) and area categories represents a sample. These are then trained using a GAE based GNN to predict the usage probabilities. In order to inform the network on the impact of the area and usage categories of the shops, they are permuted among the shops to increase the number of samples as a way of data augmentation. Such augmentation of the dataset generalize the learning process of the GNN model with ability to capture the usage probability when there will be change the shop features of the test set graphs. These are explained in detail through the following subsections. The schematic diagram shown in Figure~\ref{fig:schematic}

\subsection{Graph Extraction from Floorplans}
\begin{figure*}[htb!]
	\centering
	\begin{minipage}{.25\textwidth}
		\centering
		\includegraphics[width=\linewidth]{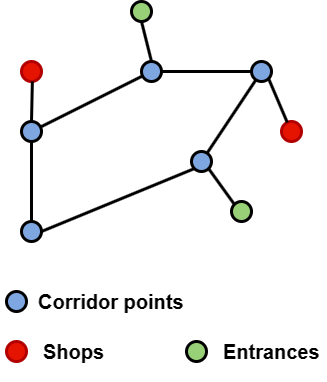}
	\end{minipage}%
	\begin{minipage}{.6\textwidth}
		\centering
		\includegraphics[width=\linewidth]{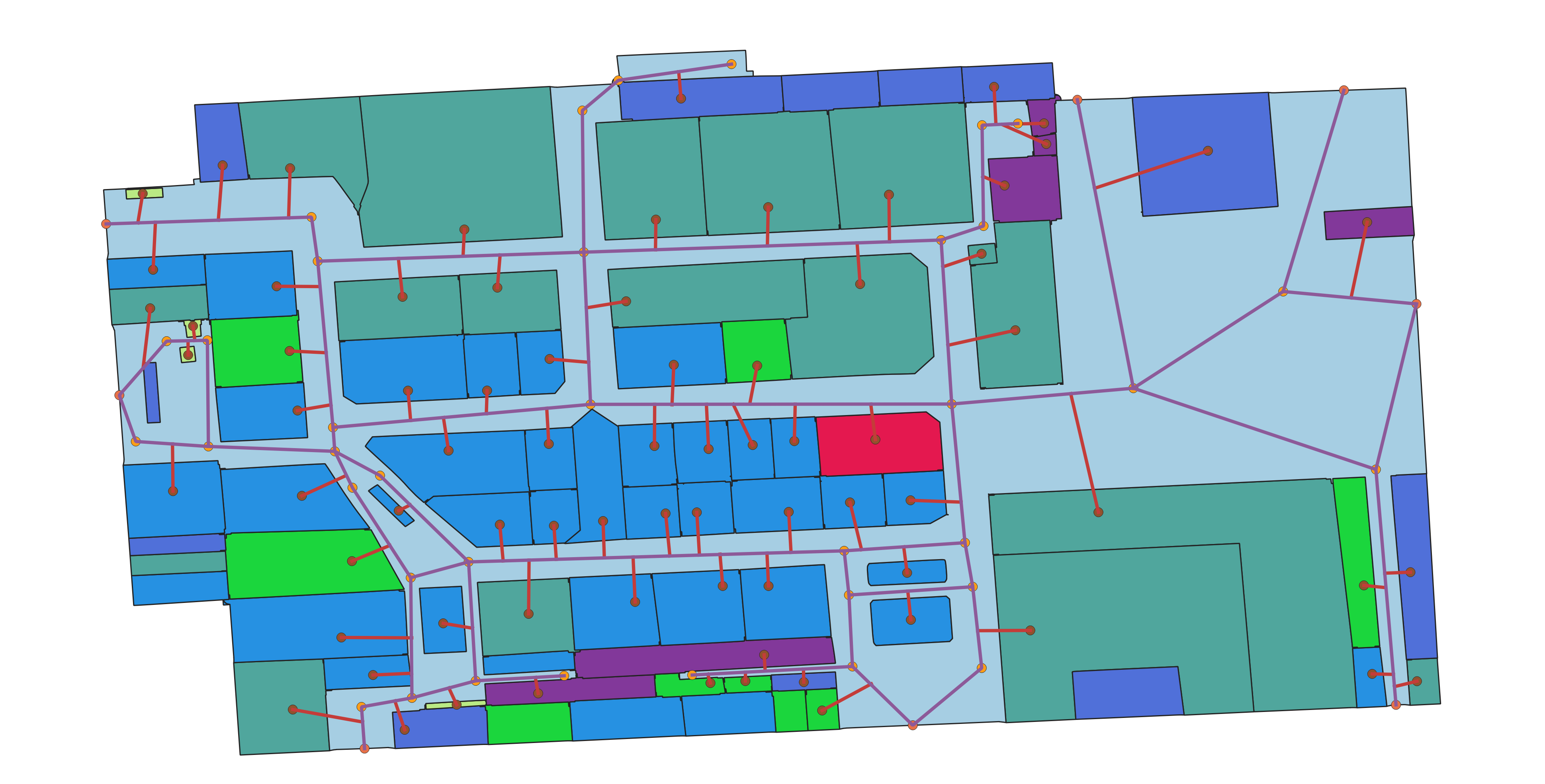}
	\end{minipage}
	\\
	\begin{minipage}{0.85\textwidth}
		\centering
		\includegraphics[width=\linewidth]{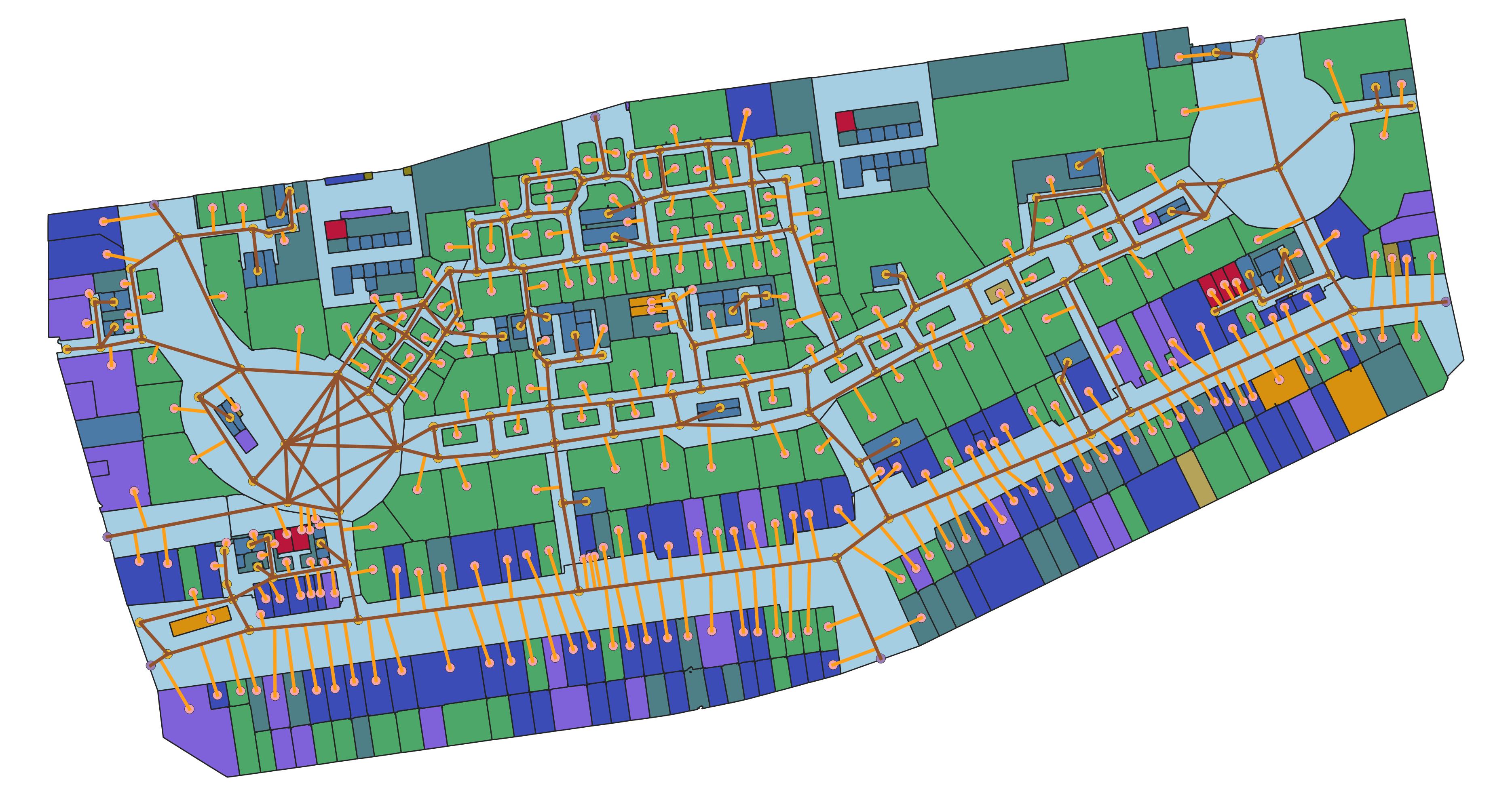}
	\end{minipage}
	\\
	\begin{minipage}{.45\textwidth}
		\centering
		\includegraphics[width=\linewidth]{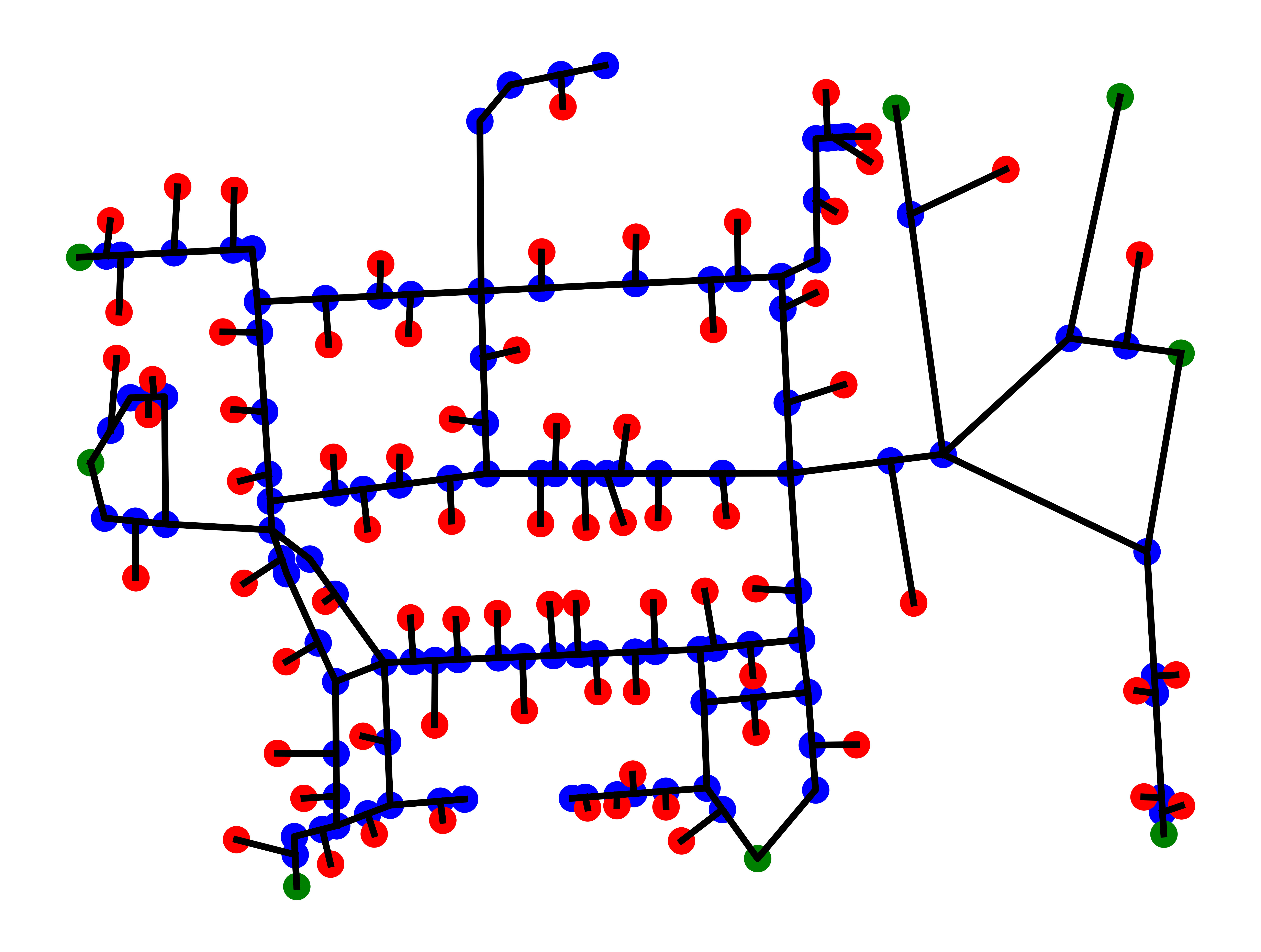}
	\end{minipage}%
	\hspace{-10pt}
	\begin{minipage}{.45\textwidth}
		\centering
		\includegraphics[width=\linewidth]{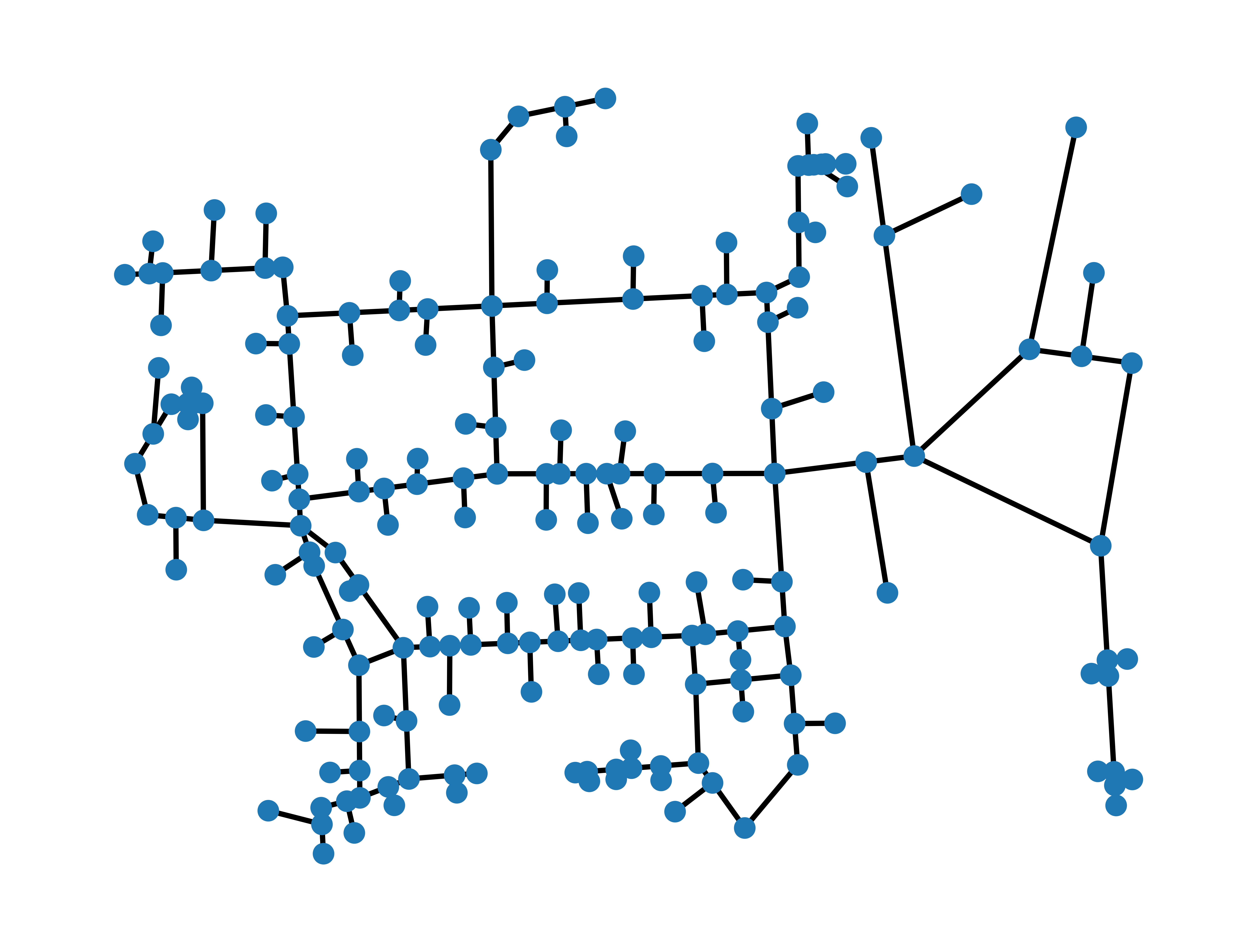}
	\end{minipage}
	\caption{Graph representation of shopping malls: (top left) A schematic graph showing different types of nodes; (top right) The extracted graph superimposed on the floorplan of Kushishang mall; (middle row) Same as top-left panel, but for Wanda Tongzhou mall; (bottom left) Graph of Kushishang mall extracted and plotted using matplotlib; (bottom right) The same visualized using NetworkX. The polygons in light blue color in the panels in the middle row and in the top right  represent corridor paths.}
	\label{fig:graphsample}
\end{figure*}

The graphs with their edges following the corridor path are obtained from a sample set of 66 floorplans of shopping malls in China. These floorplans have been obtained from the authors of reference~\cite{yang2023representation} in a vector format of \emph{ESRI shape files}, on which we have made some sparingly little modification to reclassify some polygons into categories that represent corridor paths. These polygons, which looked like corridor path, were classified as shops in the original datasets. This modification allowed most of the polygons representing corridor path form a topologically connected region as shown in light blue colour in the top-right panel of Figure~\ref{fig:graphsample}. 
As shown in the middle-row panel of Figure~\ref{fig:graphsample}, some corridor path were left to remain unconnected to the main corridors, since they looked like access paths from exclusive entrances to the building's office towers.

Figure~\ref{fig:graphsample} (top left) shows a schematic diagram of a graph and the nodes of different types. These tree types of nodes are considered during the synthesis of target data as described in the later section, however, the entrances and corridor points are treated as a single type of nodes named \emph{non-shop} in the heterogeneous-GNN based predictions.

The graph-extraction procedure follows the sequential steps as outlined below. 

Firstly, the shape file of a chosen mall's first floor's floorplan is opened in a GIS tool such as QGIS. Then two sets of point geometries have been created to represent the entrances and the corridor points on two different vector layers. Creating these points on separate layers helps in identifying the entrance nodes when computing the shortest paths from these entrances to the shops. A third vector layer of line geometry is created to connect these points as shown in Figure~\ref{fig:graphsample}(top right). Then a point layer for shops are created. These are marked inside the shop polygons, but closer towards the neighbouring corridor path. The shop nodes are connected to these corridor paths on a fifth vector layer.  

These vector layers are processed in a Python environment using tools such as Geopandas, Shapely and Numpy. The Numpy version is useful for assimilating them as graphs in PyTorch Geometric (PyG) library for GNN algorithms.~\cite{fey2019fast}  It also can be read into a Python based library NetworkX which has several shortest-path algorithms which are used during target-data synthesis described later. The visualization from Numpy/Matplotlib and NetworkX (using its tool {\tt nx.draw}) are shown in Figure~\ref{fig:graphsample} panels on bottom-left and bottom-right, respectively. These figures shows that the created shape files have been succesfully assimilated in formats required for the graph based algorithms implemented in PyG and NetworkX.

We would like to highlight that these graphs are used as a sample of an arbitrary, hypothetical or a possible mall rather than the mall from which these graphs have been extracted. It should be noted that the received floorplans of these malls in China did not contain the information on the entrances to the shops. It has been assumed in this work that each of these shops is accessed from a single corridor path. Further, these graphs are extracted only for the first level of the mall. In reality, floors are connected by a vertical traffic, which are neglected in this work. As stated earlier, the goal of this work is that the usage probabilities of different sections of the corridors can be predicted by a GNN when the shop features and corridor connections are given. These extracted graphs that is hypothetical, but yet closer to the reality of a shopping mall is useful towards this set goal. Applying the developed GNN architecture to the more accurate graph, where all floors are represented and connected via edges along vertical traffic, would be straight forward.

\subsection{Notations} \label{subsec:notation}
Before we proceed, we would like to introduce for clarity the convention used for symbols that represent a set, its number of elements, and its members. We use, bold-faced calligraphic fonts for sets. The notations, $\boldsymbol{\mathcal{V}^{(s)}}$, $\boldsymbol{\mathcal{V}^{(ns)}}$ and $\boldsymbol{\mathcal{V}^{(e)}}$ refer to the sets of shop, non-shop and entrance nodes, respectively. We use parentheses to describe the labels referring to node types and index index $p$ referring to the $p^{\textrm{th}}$ mall. For example, the notation, $\boldsymbol{\mathcal{V}^{(s,p)}}$ refer to the set of shop nodes of the $p^{\textrm{th}}$ mall. Most of the times, this index $p$ for the mall is suppressed for brevity.

The indices referring to subsets and members of a set are represented without parentheses. The indices referring to the subsets are used in superscripts (without parentheses). For example, $\boldsymbol{\mathcal{V}^{(s)ij}}$ is a subset of shop nodes with the area and usage categories being $i$ and $j$, respectively. 

The number of elements is represented by the same calligraphic font, but without being bold-faced. The member of the set is given by the same letter used for the set name added with a suffix for the index, but without using bold-face and calligraphic version of the letter. For example, the symbols $\boldsymbol{\mathcal{V}^{(s)}}$, $\mathcal{V}^{(s)}$ and $v^{(s)}_i$ represents the set of shop nodes, its number of elements, and the shop node given by the $i^{\textrm{th}}$ element in the set $\boldsymbol{\mathcal{V}^{(s)}}$. 

The total number of edges is represented by $\mathcal{E}$. The set of edges between shop and non-shop nodes is given by, $\boldsymbol{\mathcal{E}}^{(s-ns)}$. Similarly, $\boldsymbol{\mathcal{E}}^{(ns-ns)}$ is the set of edges connecting two nodes of type non-shop. Their respective adjacency matrices are given by $A^{(s-ns)}$ and $A^{(ns-ns)}$.

The feature vectors of $k^{\textrm{th}}$ shop and non-shop node of $p^{\textrm{th}}$ mall are given by $\boldsymbol{F_k}^{(s,p)}$ and $\boldsymbol{F_k}^{(ns,p)}$, respectively.

The graph of $p^\textrm{th}$ mall is represented by the notation $g^{(p)}$, or just by the letter $g$ when the index $p$ of the mall is irrelevant. The graph level features are given by the vector $\boldsymbol{F}^{(g,p)}$.

The target value, i.e., the usage probability a section of the corridor represented by $k^{\textrm{th}}$ edge in the graph is given by the notation $t_k$.

\subsection{Features for Shop Nodes}
Since our prediction of the usage probabilities of the corridor paths will be based on the features of the graph nodes, the input graphs need to be populated with proper features that describe each of those nodes accurately. It should be noted that the learnable parameters of a GNN model depends on the size of these features for each type of nodes. Therefore, it is worth consider all possible features that could differentiate and describe these nodes. An increase in the number of features is the main avenue in GNN models to increase the learnable parameters given that the other avenue of increasing the number of layers results in the well-known over-smoothing issue.

For the ease of description, let $\mathcal{V}$ be the total number of nodes of both types (i.e., the shop and non-shop nodes) in the graph. Recall from the previous section that $\mathcal{V}^{(s)}$, $\mathcal{V}^{(ns)}$ and $\mathcal{V}^{(e)}$ are the number of shop, non-shop and entrance nodes, respectively. Therefore, $\mathcal{V} = \mathcal{V}^{(s)} + \mathcal{V}^{(ns)}$, and the number of non-entrance nodes on the corridors are given by $\mathcal{V}^{(ns)}-\mathcal{V}^{(e)}$. 

We categorize the areas of the shops into 5 classes, namely, $A_i$ with $i \in \{1, \cdots, 5\}$. These are ordered such that $A_1$ refers to the category with lowest areas, and $A_5$ refers to the category with the highest area in the mall. These categories are 5 bins obtained from the range of maximum and minimum of the areas.

We also consider a similar number of classes for their level of service (Los), which is a measure of usage per area of the shops. The classes for this attribute are named $U_i$ with $i \in \{1, \cdots, 5\}$. Same as to $A_i$'s, the $U_i$'s too are ordered such that $U_1$ refers to the lowest usage category, and the $U_2$ refers to the highest. Though these categorizations are specific to each mall, one could anticipate a pattern common among all the malls. For example, a convenience stores or a supermarket is expected to have a high Los, whereas a high-end luxury boutique stores are expected to be having a low usage density with reduced footfalls.  

These two attributes, $A_i$ and $U_i$ form a feature vector of size $10$ through one-hot coding for each of the shop nodes. The first five entries of this vector denotes the class number $i$ of area, and the rest of the vector signifies the class number of the Los.

The feature vectors for the non-shop nodes are engineered as will be described in a latter section after describing how the target values are prepared from a probability model, which is described in the following.

\subsection{Probability Model}
The target data, i.e., the PoU on every edge of the graph is obtained from a probabilistic pedestrian-routing problem. 

It is evident from the definition of the usage density that shops with lower usage density would attract a smaller number of people when their areas are similar. Similarly, when the Los is same between two shops, the shop with lower area would attract a lesser number of people than the other that has a higher area. Let $p_i^{(a)}$ and $p_j^{(u)}$ represent probabilities of attraction due to the area category $A_i$ and the shop-usage category $U_j$, respectively. Naturally, we have $p_i^{(a)}<  p_j^{(a)}$ for $i<j$, and a similar behaviour for $p_i^{(u)}$. This monotonicity of  $p_i^{(a)}$ and  $p_j^{(u)}$ with respect to the class indices $i$ and $j$ suggest that, as a first approximation, $p_i^{(a)}$ and  $p_i^{(u)}$ can be modelled as straight lines of positive slope with appropriate normalizations. This results in we using the following definitions for these probabilities:
\begin{align}
	p_i^{(a)} &= \frac{1+m_a i}{5+15m_a} \ \ \ \ \mbox{and} \label{pia}\\
	p_j^{(u)} &= \frac{1+m_u j}{5+15m_u}, \label{pju}
\end{align}
where, $m_a$ and $m_u$ are two positive constants that could be fixed by field observations. In this paper, these constants are fixed as $m_a = 1$ and $m_u = 0.5$. The Equations~(\ref{pia})--(\ref{pju}) follow the required normalization, $\sum_{i=1}^5p_i^{(a)} = \sum_{j=1}^5p_j^{(u)} = 1$. From these relations, the combined probability of attraction of a shop node with area category $A_i$, and the usage-density category $U_j$ is given by,
\begin{equation}
	p_{ij} = p_i^{(a)}p_j^{(u)} \label{pij}.
\end{equation}
Hence, $\sum_{i,j}p_{ij} = 1$. Since every shop node in the graph has these categories as their features, each of them are assigned with an attraction probability given by Equation~(\ref{pij}).

Though the GNN model that we will present in a latter section is more general in a way to predict the usage probabilities of different sections of the corridor paths based on the shop features and the underlying graph, we would like to restrict the training datasets such that the number of shops in each of the categories $A_i$ and $U_i$ follow some rational expectations in realistic malls. 

For the purpose of clarity, let $\hat{n}(A_i)$ be the number of shop nodes with the area-category $A_i$, and a similar definition for $\hat{n}(U_j)$. We expect the following two patterns in a typical shopping mall:
\begin{align}
	& \hat{n}(A_1) < \hat{n}(A_2) < \hat{n}(A_3) > \hat{n}(A_4) > \hat{n}(A_5), \ \mbox{and} \label{Ai_convex}\\
	& \hat{n}(U_1) < \hat{n}(U_2) < \hat{n}(U_3) > \hat{n}(U_4) > \hat{n}(U_5). \label{Uj_convex}
\end{align}
The pattern in the Inequality~(\ref{Ai_convex}) suggest that the number of shops with medium area is greater than that of other areas, and that the number of shops decreases when the category-number $i$ increase or decrease from $i=3$. This means that the shops as small as a bodega or those with areas as large as a supermarket will be fewer than the other shops of medium size.

A similar distribution of the shops is expected with respect to the usage density as shown in the Inequality~(\ref{Uj_convex}) in order to optimize between the footfalls and shopping experience. 

These Inequalities~(\ref{Ai_convex})--(\ref{Uj_convex}) can be modelled as a Gaussian distributions $f_a$ and $f_b$ centered around $A_3$ and $U_3$, i.e., 
\begin{align}
	f_a &= \mathcal{Z}(3,\sigma_a) \ \ \ \mbox{and} \label{eq:fa}\\
	f_u &= \mathcal{Z}(3,\sigma_u), \label{eq:fu}
\end{align}
where the function $\mathcal{Z}$ stands for the Gaussian distribution, $\mathcal{Z}(\mu,\sigma) = (\sigma\sqrt{2\pi})^{-1}\exp[-(x-\mu)^2/(2\sigma^2)]$ for the variable $x$. The standard deviations $\sigma_a$ and $\sigma_b$ could be estimated from field observations as for the constants $m_a$ and $m_u$ in Equations~(\ref{pia})--(\ref{pju}). However, for the purpose of demonstration of this strategy fix these $\sigma$'s as $\sigma_a = \sigma_u = 1.1$. The bins are considered as follows: $\{x<1.5\}\in A_1$, $\{1.5\leq x<2.5\}\in A_2$, $\{2.5\leq x <3.5\}\in A_3$, $\{3.5< x\leq4.5\}\in A_4$, and $\{x>4.5\}\in A_5$. The bins with respect to $U_j$ is also considered same in this work without loss of generality.

\subsection{Assignment of Shop-Node Features}
The application of the distributions in Equations~(\ref{eq:fa})--(\ref{eq:fu}) on the $\mathcal{V}^{(s)}$ number of shop nodes results in $\mathcal{V}^{(s)ij}$ number of shop nodes with area and usage density attributes as $A_i$ and $U_j$ for each $i\in \{1,\cdots,5\}$ and $j\in \{1,\cdots,5\}$. Therefore $\sum_{i,j}\mathcal{V}^{(s)ij} = \mathcal{V}_s$. 
\begin{algorithm}
	\caption{Shop-node feature assignments. The notation, $A\setminus B$ means the subtraction of the set $B$ from set $A$.}
	\begin{algorithmic}
		\State $M \gets$ \textbf{No. of Malls}
		\State $S \gets$ \textbf{No. of samples for each mall}
		\Repeat
		\For {$p\leftarrow 1, M$}
		\State $g_p \gets$ \textbf{featureless graph of mall} $p$
		\State $\mathcal{V}^{(s,p)} \gets$ \textbf{No. of shop nodes in} $g_p$
		\For {$q\leftarrow 1, S$}
		\State $\boldsymbol{\mathcal{V}}^{(s,p)} \gets$ \textbf{set} $\{1,\cdots,\mathcal{V}^{(s,p)}\}$
		\For {$i\leftarrow 1, 5$}
		\For {$j\leftarrow 1, 5$}
		\State $\mathcal{V}^{(s,p)ij} \gets$ \textbf{No. of shop nodes with} \\ \hspace{4cm}\textbf{with features drawn from} \\
		\hspace{4cm}\textbf{distributions} $f_a$ \textbf{and} $f_b$
		\State $\boldsymbol{\mathcal{V}}^{(s,p)ij} \gets$ \textbf{randomly drawn} $\mathcal{V}^{(s,p)ij}$ \\
		\hspace{4cm} \textbf{samples from} $\boldsymbol{\mathcal{V}}^{(s,p)}$
		\State $\boldsymbol{\mathcal{V}}^{(s,p)} \gets \boldsymbol{\mathcal{V}}^{(s,p)} \setminus \boldsymbol{\mathcal{V}}^{(s,p)ij}$
		\For {$k \in \boldsymbol{\mathcal{V}}^{(s,p)ij}$}
		\State $F_{k,l}^{(s,p)} \gets [\delta_{il}, \delta_{jl}],  \ \ \ l \in \{1,\cdots,5\}$
		\EndFor
		\EndFor
		\EndFor
		\EndFor
		\EndFor
		\Until{ \textbf{completed}} 
	\end{algorithmic}
\end{algorithm}
For each mall identified with index $p$, we consider $S$ number of samples. In this work, we set $S=200$. The index $p$ is such that $p\in\{1,\cdots, M\}$ where $M$ is the No. of malls. Since $\mathcal{V}^{(s)}$ vary among the malls, let us label $\mathcal{V}^{(s)}$ and $\mathcal{V}^{(s)ij}$ as $\mathcal{V}^{(s,p)}$ and $\mathcal{V}^{(s,p)ij}$, respectively. Let $g_p$ be the featureless graph of a mall (given by the index $p$) that defines only the nodes of both types and the edges. (However, after completion of the tasks described in this section, $g_p$ will contain features defined at node and graph level.)

In each sample, the shop nodes have distinct arrangement of their features. Let variable $q\in\{1,\cdots,S\}$ is the index of samples, and let $g_{p,q}$ represents the graph of a chosen mall labeled $p$ and the sample $q$.
For a chosen graph $g_{p,q}$, for each pairs of $(i,j)$, a set of $\mathcal{V}^{(s,p)ij}$ number of shop nodes are uniform-randomly chosen without replacement, and the features $(A_i, U_j)$ are assigned in one-hot encoded form. Therefore, the feature vector of all shop-nodes will have a length of 10. Precisely, for the $k^\textrm{th}$-shop-node with features $(A_i, U_j)$, the feature vector is given by,
\begin{equation}
	F_{k,l}^{(s,p)} = [\delta_{il}, \delta_{jl}], \ \ \ k \in \{1,\cdots,\mathcal{V}^{(s,p)}\}, \ \ \ 
	l \in \{1,\cdots,5\}
	\label{eqn:fklsp}
\end{equation}
where the symbol $[,]$ refers to horizontal concatenation and $\delta_{il}$ is Kronecker delta.

This procedure is repeated for each sample $q$ and each shopping mall $p$. This is explained in  Algorithm 1.

It should be noted that the change in the arrangement of shop attributes when the index $q$ is changed is crucial to inform the GNN later about the target variable's dependency on these attributes. One can also view each sample $q$ for the same graph of a mall $p$ as data augmentation similar to that provided by affine transformations (i.e., rotations, scalings and jittering) in the case of image based machine-learning algorithms.

\begin{figure*}[htb!]
	\centering
	\setlength{\unitlength}{0.1\textwidth}
	\begin{minipage}{0.45\textwidth}
		\centering
		\begin{picture}(4.5,4.0)
		\includegraphics[width=\linewidth]{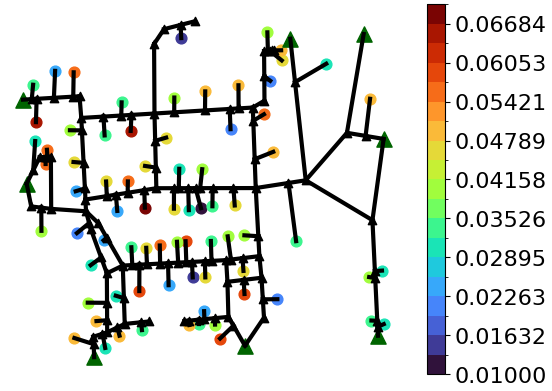}
		\put(-4.4,3.2){\large Sample 1: Probability of Attraction}
		\end{picture}
	\end{minipage}%
	\begin{minipage}{0.45\textwidth}
		\centering
		\begin{picture}(4.5,4.0)
		\includegraphics[width=\linewidth]{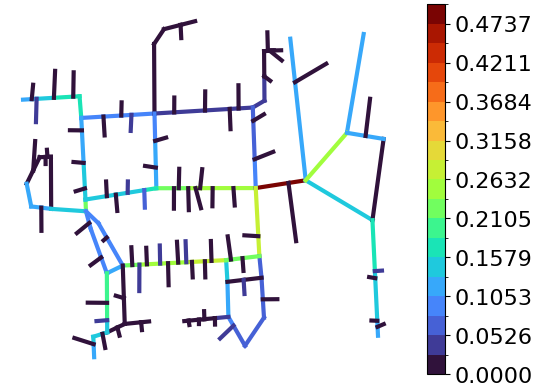}
		\put(-4.4,3.2){\large Sample 1: Usage Probability}
		\end{picture}
	\end{minipage}
	\\
	\begin{minipage}{0.45\textwidth}
		\centering
		\begin{picture}(4.5,4.0)
		\includegraphics[width=\linewidth]{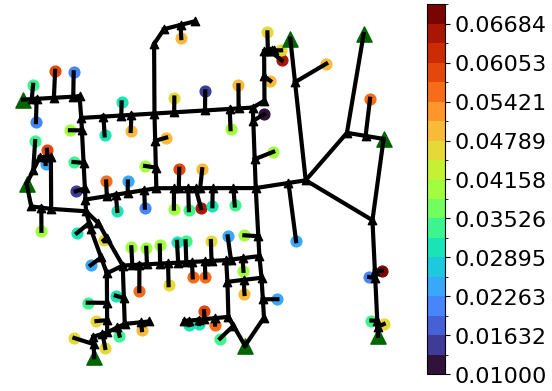}
		\put(-4.4,3.2){\large Sample 2: Probability of Attraction}
	    \end{picture}
	\end{minipage}%
	\begin{minipage}{0.45\textwidth}
		\centering
		\begin{picture}(4.5,4.0)
		\includegraphics[width=\linewidth]{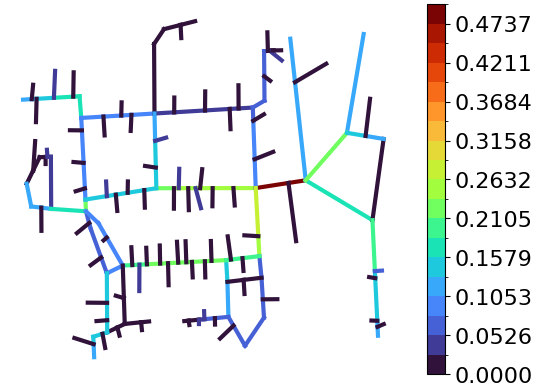}
		\put(-4.4,3.2){\large Sample 2: Usage Probability}
	    \end{picture}
	\end{minipage}
	\caption{The attraction and usage probabilities: (Left column) The attraction probabilities of the shop nodes of Kushishang mall for two samples in the training dataset. The shop nodes are shown as coloured spots. The non-shop nodes have been shown using triangle shaped markers. The black triangles refer to the nodes on the corridor paths, while the dark-green triangles refer to the entrance nodes. (Right column) the corresponding usage probabilities of graph edges.}
	\label{fig:attr_prob_and_target}
\end{figure*}

Once the shop-node features are assigned, each of them have a properly defined attraction-probabilities given by Equation~(\ref{pij}). The Figure~\ref{fig:attr_prob_and_target}(left) shows them for a sample of feature assignment to the shop nodes.

\subsection{Target Data Preparation: Usage Probabilities for Graph Edges}
Let $\mathcal{E}$ be the total number of all types of edges in the graph. The target data, i.e., the usage probabilities, $t_k$ for each $k \in \{1,\cdots,\mathcal{E}\}$ is obtained by making use of the probability of attraction, $p_{ij}$ for the shop nodes given by Equation~(\ref{pij}) and by using the \textit{Dijkstra} algorithm for shortest path. 

Since the procedure described here is common for all malls, Let is suppress the mall index, $p$ for clarity.

Let $N$ be the number of people in the mall at a point of time. Then number of people that are supposed to visit shops having the feature $(A_i, U_j)$ are given by 
\begin{equation}
	N_{ij}=p_{ij} N.
	\label{eqn:NijpN}
\end{equation} 
Let us recall that $\mathcal{V}^{(s)ij}$ (after suppressing the label $p$ for the mall) is  the number of shops having the feature $(A_i, U_j)$. Using the Dijkstra algorithm implemented in NetworX, a set of $\mathcal{V}^{(s)ij}$ shortest paths are determined from each of the entrances to all the nodes having $A_i$ and $U_j$ as area and usage categories.

The total number of people $N$ would not affect the results in theory, since we compute the probabilities of usage of each graph edge, and not the number of people using it. However, the Equation~(\ref{eqn:NijpN}) could introduce round of errors in $N_{ij}$, if $N$ is not sufficiently large. Therefore, and with another criterion described below, it is optimized for efficiency and accuracy as follows. 

As a first step, the $N$ is chosen as the smallest possible number, such that $N_{ij}$ are integers, and that there are no greatest-common divisors other than unity between all of $N_{ij}$'s. 

Then, the $N$ is multiplied by a constant integer $c_{ij}$ and set equal to $N$, i.e., $N \leftarrow c_{ij} N$ such that $N_{ij}$ becomes divisible by $\mathcal{V}^{(s)ij}$’s for all $i$'s and $j$'s. The latter step is needed due to the fact that we need to loop through all $\mathcal{V}^{(s)ij}$ number of nodes when we compute the shortest path from all entrances. Therefore we can allot $N_{ij}/\mathcal{V}^{(s)ij}$ number of people for each shop having the attributes $(\mathcal{A}_i, \mathcal{U}_j)$. Then this number $N$ is multiplied by the total number of entrances to the mall $\mathcal{V}^{(\mathrm{ent})}$. Finally, The number of shortest paths running through each of the edges in the graph are counted and divided by $N$. 

This can be described through a mathematical expression as follows. Following the convention used in this paper, let $\boldsymbol{\mathcal{V}}^{(s)ij}$ represent the set of $\mathcal{V}^{(s)ij}$ number of shop nodes each having the attributes $(A_i,U_j)$. Let $\mathcal{X}$ be the set of entrance nodes. Then for each edge $e_k$, the usage probability $t_k$, i.e., the target data, is given by,

\begin{equation}
	t_k = \frac{1}{N}\sum_{v^{(e)}\in \mathcal{V}^{(e)}} \sum_{i=1}^5 \sum_{i=1}^5 \sum_{v^{(s)}\in \boldsymbol{\mathcal{V}}^{(s)ij}} \eta(v^{(e)},v^{(s)}|k), \ \ \ \ k\in\{ 1,\cdots,\mathcal{E} \},
	\label{eqn:tk}
\end{equation}
where $\eta(v_x,v_s|k)$ is the total number of people who crossed the graph edge $e_k$ by starting from the entrance node $v_x$ to the shop node $v_s$ via the shortest path. This number is equal to the number, $N_{ij}/\mathcal{V}^{(s)ij}$ for all edges on the route from $v_x$ and $v_s$. Since, $t_k$ given by the equation is the probability of the usage of each edge which can appear in multiple shortest paths, $\sum_{k=0}^\mathcal{E}t_k=1$ would not hold.

The obtained usage probabilities of two samples of Kushishang mall is shown in Figure~\ref{fig:attr_prob_and_target}(right column) corresponding to the feature assignment that gave rise to the attraction probabilities shown in Figure~\ref{fig:attr_prob_and_target}(left column). We verified that the sum of the usage probabilities of the edges connecting to each of the entrance nodes (shown in dark-green triangles in Figure~\ref{fig:attr_prob_and_target}(left column)) are having the same value of $t_k$'s. This is observable to the naked eye on the Figure~\ref{fig:attr_prob_and_target}(right column) on the entrance nodes that have single degree. This equality in $t_k$'s arise since all entrance nodes have been treated to have equal probability for a shopper to choose. 

However some entrances depending on the proximity to other amenities outside the mall may be more preferable in some realistic situation. In such cases, the implementation of the algorithm described in this section could be modified accordingly. Especially, the expression $\sum_{v^{(x)}\in \mathcal{X}}$ in Equation~(\ref{eqn:tk}) will need to be modified in such situation by drawing the $v^{(x)}$ from the sample space of the entrance nodes $\mathcal{X}$ satisfying the probabilities prescribed for each of them. 

The highest usage probability in Figure~\ref{fig:attr_prob_and_target}(right column) is found for an edge (shown in dark red colour) that bridges too groups of nodes on either sides, hence, this is understandable exactly due to this bridge like behaviour. 

When comparing the usage probabilities between the two samples shown in Figure~\ref{fig:attr_prob_and_target}(right column), we observed that the $t_k$'s largely similar on the main corridors. However, they differ greatly on the edges connecting the main corridors to their respective shops. The apparent insensitivity of the $t_k$'s on the main corridor to the attraction probabilities of the shops in this particular choices of samples is due the fact that these attraction probabilities seen on the Figure~\ref{fig:attr_prob_and_target}(left column) are well shuffled, and that the number shops having similar values of these attractions on a corridor section do not vary drastically. This would not be the case when the shop-nodes' attraction probabilities are inhomogeneously distributed with visible spatial gradients in the 2D plane of the figure.

\subsection{Comparing Usage Probabilities with Graph Centrality Measure}

The usage probabilities given by $t_k$'s in Equation~(\ref{eqn:tk}) is more appropriate than that could be obtained from other measures from graph theory. For example, the \emph{betweenness centrality} of the graph assigns the score for each edge purely based on shortest paths that passes through them disregard of the node types and node features. It is worthwhile to compare our usage probabilities with the betweenness centrality measure.

The betweenness centrality, $c_k$ of a any edge with id $k$ is defined as
\begin{equation}
	c_k = \frac{2}{\mathcal{V}(\mathcal{V}-1)}\sum_{j>i}^{\mathcal{V}} \sum_{i=1}^{\mathcal{V}} \zeta(i,j;k), \ \ \ \ k\in\{ 1,\cdots,\mathcal{E} \}, \label{eqn:ck}
\end{equation}
where $\zeta(i,j;k)$ is the number of shortest paths from a node with an id $i$ to another node with an id $j$ that passes through the edge with id $k$. The measure $c_k$ is relevant to compare and against the results that we obtained for $t_k$'s since both of them involves shortest path between nodes.
\begin{figure*}[htb!]
	\centering
	\vspace{0pt}
	\setlength{\unitlength}{0.08\textwidth}
	\begin{picture}(10,2.5)
	\begin{minipage}{.4\textwidth}
		\centering
		\includegraphics[width=\linewidth]{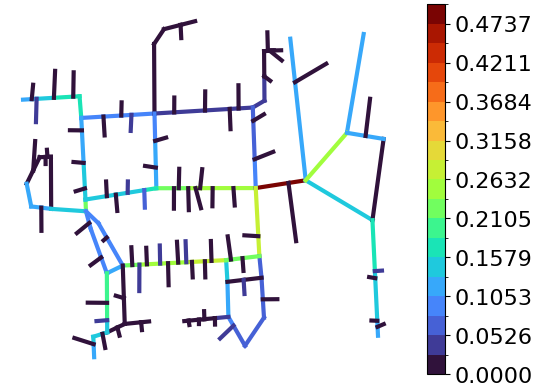}
	\end{minipage}%
	\hspace{0pt}
	\begin{minipage}{.4\textwidth}
		\centering
		\includegraphics[width=\linewidth]{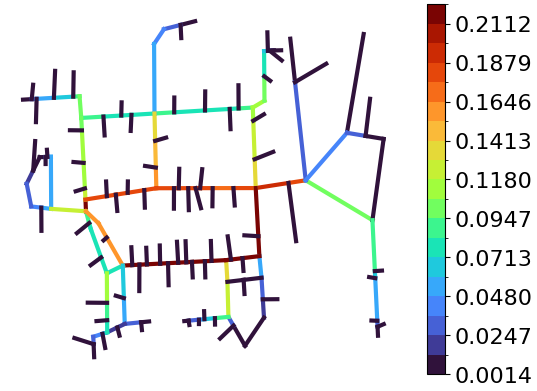}
	\end{minipage}
	\put(-6.5,2.0){\large Kushishang Mall}
	\end{picture}
	\vspace{3cm}
	\\
	\vspace{0pt}
	\begin{picture}(10,2.5)
	\begin{minipage}{.4\textwidth}
		\centering
		\includegraphics[width=\linewidth]{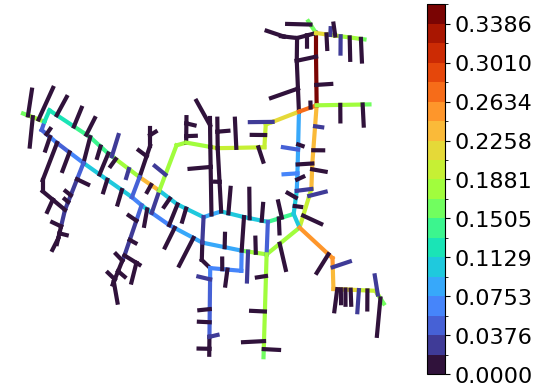}
	\end{minipage}%
	\hspace{0pt}
	\begin{minipage}{.4\textwidth}
		\centering
		\includegraphics[width=\linewidth]{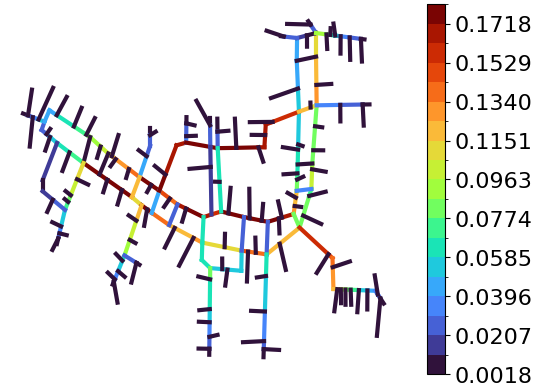}
	\end{minipage}
	\put(-6.5,2.0){\large Wanda Fengkedian Mall}
	\end{picture}
	\vspace{3cm}
	\\
	\vspace{0pt}
	\begin{picture}(10,2.5)
	\begin{minipage}{.4\textwidth}
		\centering
		\includegraphics[width=\linewidth]{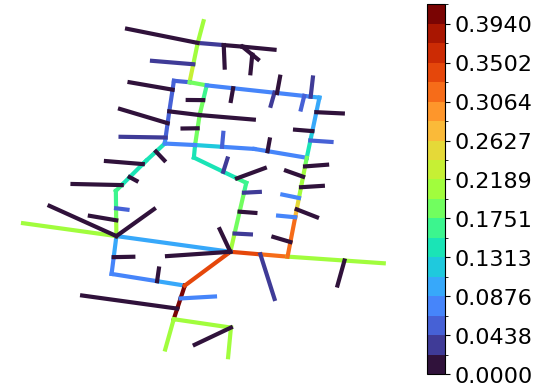}
	\end{minipage}%
	\hspace{0pt}
	\begin{minipage}{.4\textwidth}
		\centering
		\includegraphics[width=\linewidth]{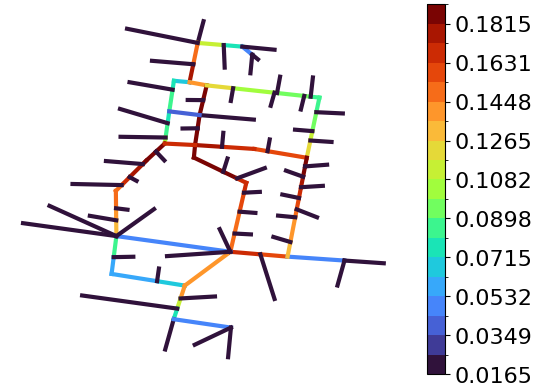}
	\end{minipage}
	\put(-6.5,2.0){\large Yintaibaihuo DHM Mall}
	\end{picture}
	\vspace{2.5cm}
	\caption{The usage probabilities and the betweenness centrality of the graph edges of three malls, namely, Kushishang, Wanda Fenkedian and Yintaibaihuo DHM. (left column) The usage probabilities. (right column) The betweenness centrality.}
	\label{fig:bet_cent}
\end{figure*}
The right column of Figure~\ref{fig:bet_cent} shows these $c_k$'s from Equation~(\ref{eqn:ck}) for each of the edges of the graphs of three malls in Beijing with names Kushishang (top row), Wanda Fengkedian (middle row) and Yintaibaihuo DHM (bottom row). The corresponding $t_k$'s obtained from Equation~(\ref{eqn:tk}) for a sample from each mall have been shown on the left column of the same figure for comparison. Similar to $t_k$'s in Figure~\ref{fig:bet_cent}(left column), the $c_k$'s of edges shown in Figure~\ref{fig:bet_cent}(right column) that leads to single degree nodes in each graph have the lowest values, as can be expected. 

However, the rest of the characteristics of the distribution of $t_k$'s are in stark contrast with that of $c_k$'s. As can be seen from this figure, the $c_k$'s are highest for the edges that are well in the interior portions of the graphs than those edges where $t_k$'s are higher. 

We observe that the highest values of $t_k$'s are still in the interior region, however, shifted towards one of the entrance nodes in each graph. This can be seen in the case of Kushishang mall in Figure~\ref{fig:attr_prob_and_target} where the entrance nodes have been shown on the left panel as dark-green triangles.

The $t_k$'s and $c_k$'s differ in their distribution due to two factors. Firstly, the $t_k$'s consider only the shortest paths between an entrance node and a shop node, whereas the general definition of the betweenness centrality $c_k$ that is commonly used in graph analysis as in Equation~(\ref{eqn:ck}) count that are between all types of nodes, including between two non-shop nodes in the corridors. 

Secondly, the $t_k$ of a graph is a function of the features of the destination shop nodes unlike the $c_k$. Though this is not explicitly obvious from Equation~(\ref{eqn:tk}), it should be noted that the number people visiting the destination node, $\eta$ in that equation is obtained from the probability model described. 

In summary, the $t_k$'s are more appropriate than the centrality measures, since they are sensitive to node types and features. Their dependency on node-features is crucial to train the GNN model later in a feature-driven approach unlike in the cases where the prediction is based on time history. The dependency on the node type will be effectively handled by our GNN model later through the incorporation of heterogeneity in the architecture.

Now that we have described about the shortest path strategy for the target labels, we are in the position to describe in the next subsection how we harvest the features for the non-shop nodes.

\subsection{Features for Non-shop Nodes}
As we have described in the Introduction, we do not adopt the method of singular-value decomposition or that of the well-known \emph{node2vec} since the model training is not performed on a large single graph like that of Cora citation datasets, but on several different graphs of varying number of nodes and edges. Instead, we adopt a method that is suitable for the present context where the shortest paths and the features of the destination-shop nodes play roles.

Harvesting such features for the otherwise featureless non-shop nodes helps to give increase the learnable parameters in the weight matrix that multiplies them, besides giving the harvested information to the network.

We consider 10 features for each of the non-shop nodes. The first five features corresponds to the number of shortest paths that a particular non-shop node is part of for each of the 5 area categories of the destination shops. The next five features are defined in similar way but corresponds to the five usage-density categories of the destination-shop nodes.

More precisely, the feature vector, whose elements are given by the subscript $m$ for a non-shop node with id $l$ can be written as follows.
\begin{strip}
\begin{equation}
	F_{l,m}^{(ns,p)} = 
	\begin{cases}
		\frac{1}{\sum_{j = 1}^5 \mathcal{V}_{mj}^{(s)}}\sum_{v^{(x)} \in \mathcal{X}} \sum_{\{v^{(s)}: A(v^{(s)})=m\}} \zeta(v^{(x)},v^{(s)}|l) & \mathrm{for} \ m \in \{1,\cdots,5\}   \\
		\frac{1}{\sum_{i = 1}^5 \mathcal{V}_{im}^{(s)}}\sum_{v^{(x)} \in \mathcal{X}} \sum_{\{v^{(s)}: U(v^{(s)})=m-5\}} \zeta(v^{(x)},v^{(s)}|l) & \mathrm{for} \ m \in \{6,\cdots,10\}
	\end{cases}
	\label{eqn:flm}
\end{equation}
\end{strip}
where the functions, $A(v)$ and $U(v)$ are defined to fetch the node $v$'s the area and usage density categories, respectively, and the definition of the notation $\zeta(i,j;k)$ is same as the one given under Equation~(\ref{eqn:ck}). The Equation~(\ref{eqn:flm}) gives the features of all non-shop nodes including those on the corridor path and the entrance nodes. As in the cases of  Equations~(\ref{eqn:ck}) and~(\ref{eqn:tk}), we used the Dijkstra's algorithm implemented in NetworkX for finding the shortest paths.

\subsection{Graph-Level Features}
Across the dataset of different malls, the graphs change as expected. This suggests that there need to be features describing the global aspects of each graph. These graph-level features would help the GNN models by informing them that the graph has changed. An accurate description of these features could come from a graphlet decomposition~\cite{prvzulj2007biological, shervashidze2009efficient} of each graph. However, we consider a simple set of 32 features as a trade off with computational resource. These features for a mall with index $p$ can be written as
\begin{equation}
	\mathbf{F}^{(g,p)} = [\mathcal{V}^{(s,p)}, \mathcal{V}^{(ns,p)},\mathcal{V}^{(e,p)},\mathbf{a}, \mathbf{b}],
\end{equation} 
where the first three entries on the RHS refer to the number of shop, non-shop and entrance nodes. (Recall the notations introduced in the subsection~\ref{subsec:notation}.) As in Equation~(\ref{eqn:fklsp}), the brackets $[a,b,\cdots]$ imply the horizontal concatenation of elements inside. The row vectors, $\mathbf{a}$ and $\mathbf{b}$ each have 25 and 4 elements in them, respectively. The elements of $\mathbf{a}$ are given by 
\begin{equation}
a_m = \mathcal{V}^{(s,p)}_{1+\mathrm{floor}(m/5),\ \  1+\mathrm{mod}(m,5)}, \ \ \  m = \{1,\cdots,25\},
\end{equation}
which is simply the flattened version the matrix given by the elements, $\mathcal{V}^{(s,p)ij}$. The elements of $\mathbf{b}$ are the histogram of the number of nodes having degrees 1 to 4. 

It should be noted that these graph-level features are needed only when the dataset comprises the graphs of different malls. If the GNN training is performed on a single mall where each data corresponds to the sample, these $\mathbf{F}^{(g,p)}$ will be a set of redundant features. More on this will be briefed in the next section.

Finally, the graphs of all samples of different malls together with target values on edges, and the node- and graph-level features are saved as a Python list of the PyTorch Geometric package's \emph{Data} objects as indicated in Figure~\ref{fig:schematic}. These graphs in this list are shuffled and split into datasets for training and testing.  

\section{GNN Model and Results}
\begin{figure*}[htb!]
	\centering
	\setlength{\unitlength}{0.1\textwidth}
	\begin{tabular}{l @{\hskip 0.00\textwidth} r}
		\begin{tabular}{l}
			\begin{minipage}{0.375\textwidth}
				\centering
				\begin{picture}(3.75,8.9)
					\put(0,0){\includegraphics[width=\linewidth]{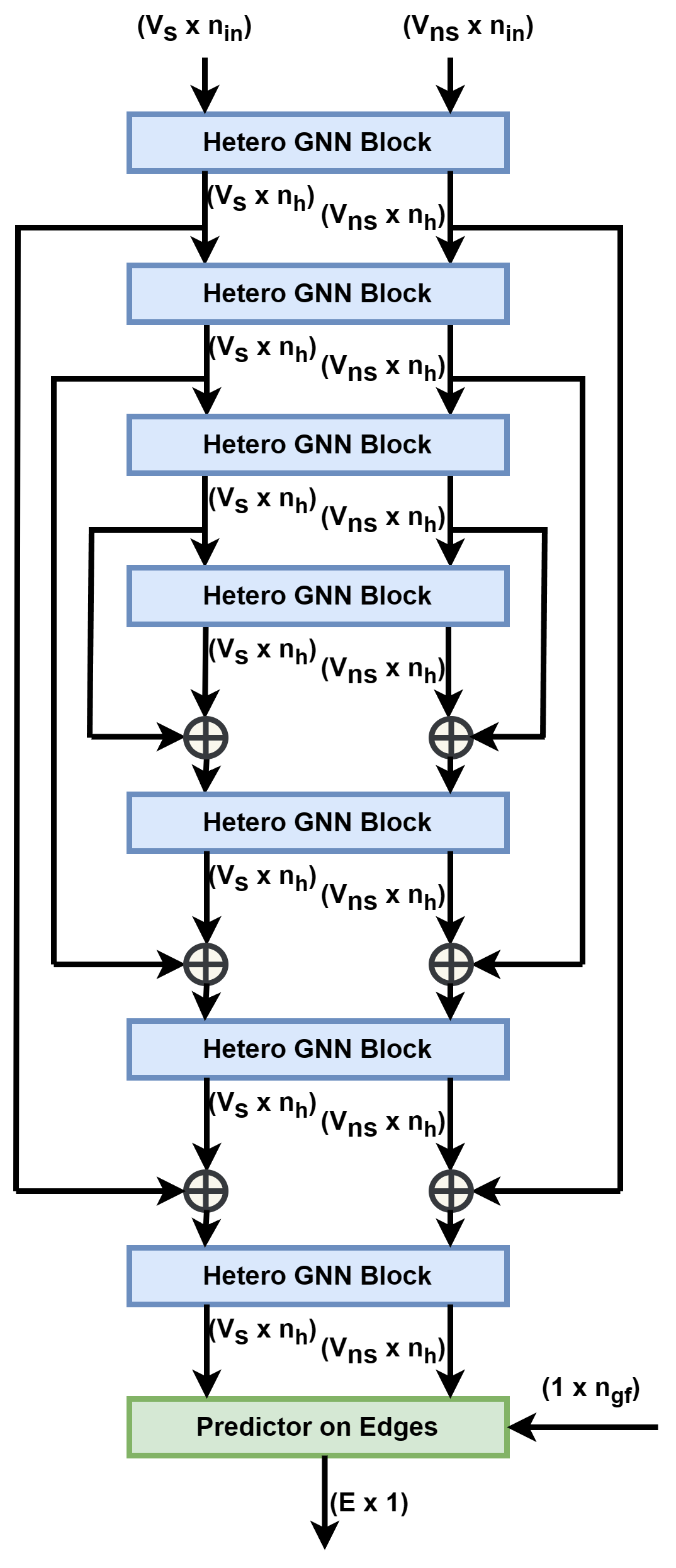}}
					\put(1.4,8.85){\parbox{3.0cm}{\bf \sffamily Main Model}}
				\end{picture}
			\end{minipage}%
			\\
			\begin{minipage}{0.375\textwidth}
				\centering
				\begin{picture}(3.75,1.8)
					\put(0,0){\includegraphics[width=\linewidth]{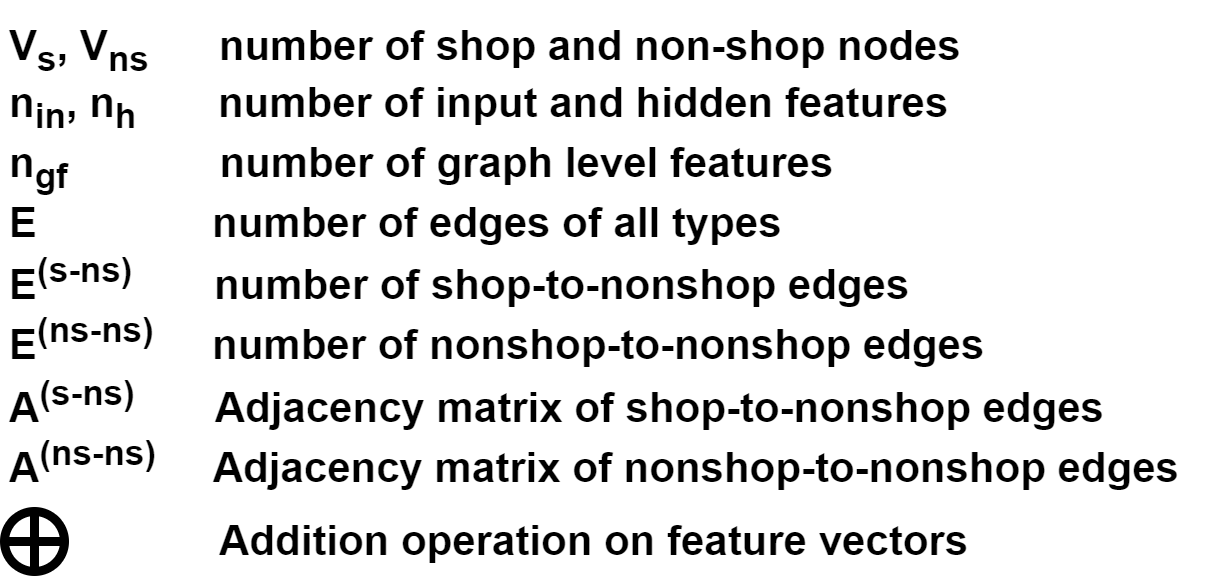}}
				\end{picture}
			\end{minipage}%
		\end{tabular}
		& 
		\begin{tabular}{c}
			\begin{minipage}{0.4\textwidth}
				\centering
				\begin{picture}(4.0,7.2)
					\put(0,0){\includegraphics[width=\linewidth]{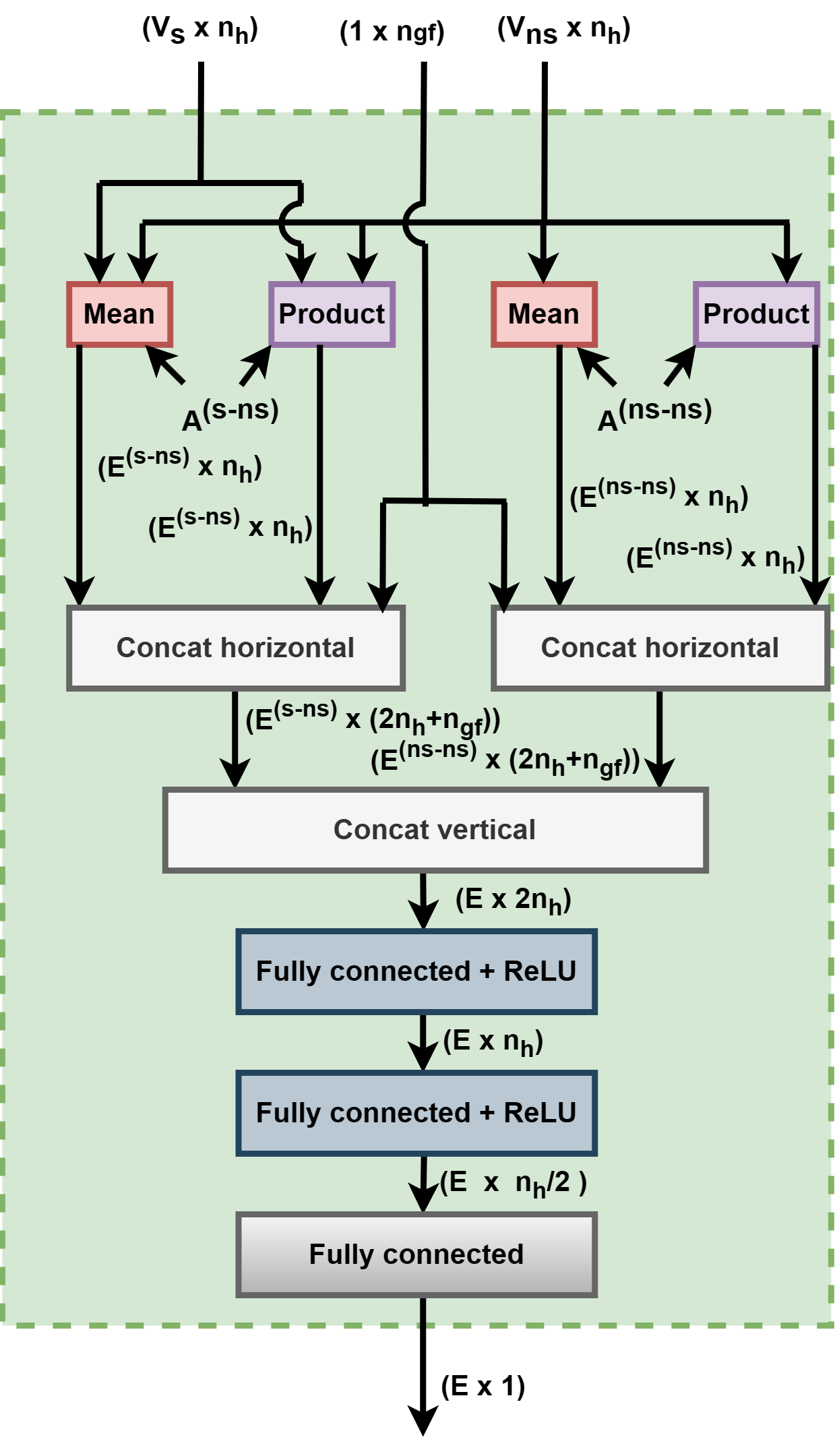}}
					\put(1.0,7.0){\bf \sffamily Predictor on Edge Block}
				\end{picture}
			\end{minipage}
			\\
			\begin{minipage}{0.35\textwidth}
				\centering
				\begin{picture}(3.5,3.8)
					\put(0,0){\includegraphics[width=\linewidth]{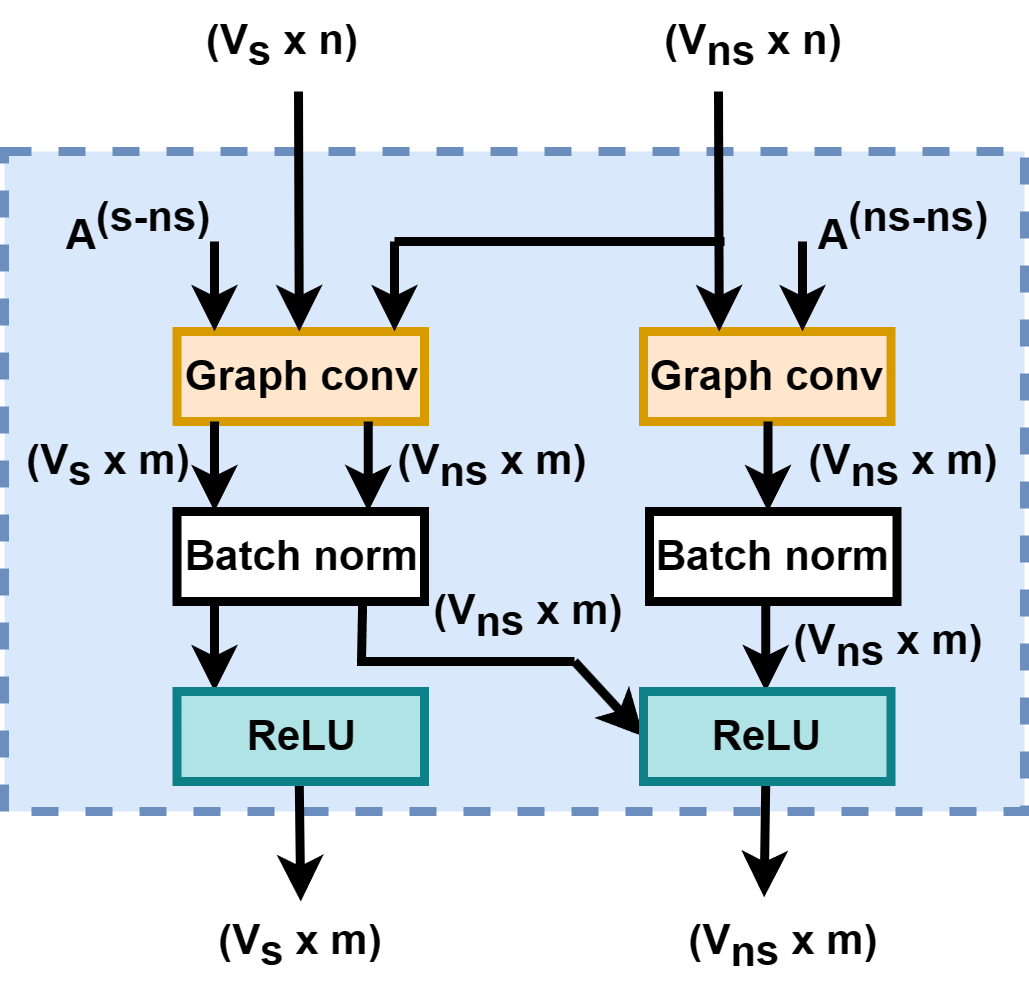}}
					\put(0.9,3.4){\parbox{4.0cm}{\bf \sffamily Hetero-GNN Block}}
				\end{picture}
			\end{minipage}
		\end{tabular}
	\end{tabular}
	\caption{The GNN-model architecture. The notations and symbols are described in the bottom left corner. Note that these are different from the notations used in the text. In this paper, $n_\textrm{in} = 10$ and $n_h = 16$}
	\label{fig:gnn_model}
\end{figure*}
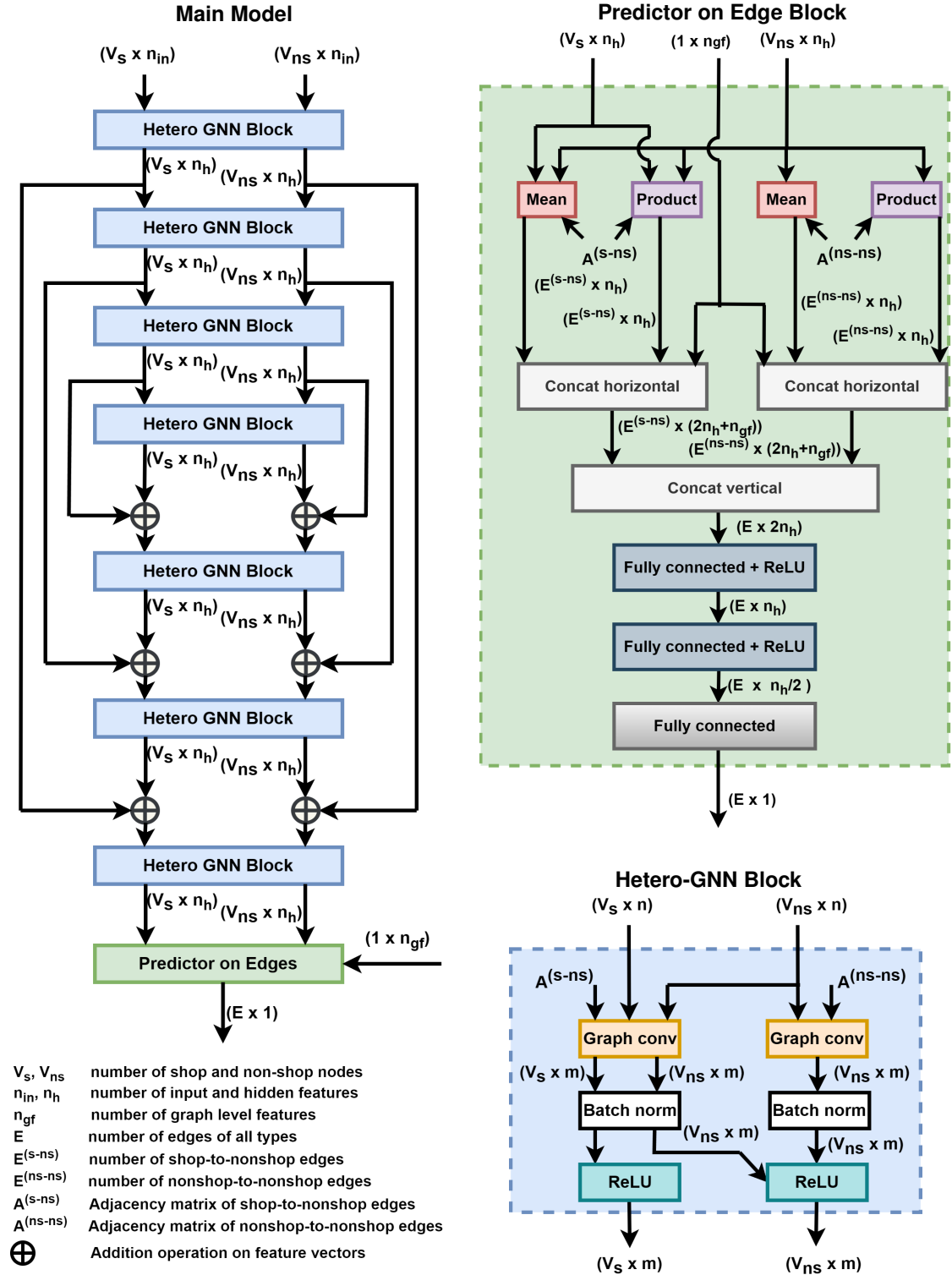

We use a heterogeneous GNN model shown in Figure~\ref{fig:gnn_model} that uses the well-known message passing layers such as, GraphSAGE, GCN, GAT and GIN with two types of nodes, namely, shop nodes and non-shop nodes. This falls under the category of graph auto-encoders(GAE), since we use message passing GNN layers for node-level encoding followed by a decoder for predicting a variable on the edges.

\begin{figure*}[htb!]
	\centering
	\vspace{0pt}
	\setlength{\unitlength}{0.08\textwidth}
	\begin{picture}(10,2.5)
		\begin{minipage}{.4\textwidth}
			\centering
			\includegraphics[width=\linewidth]{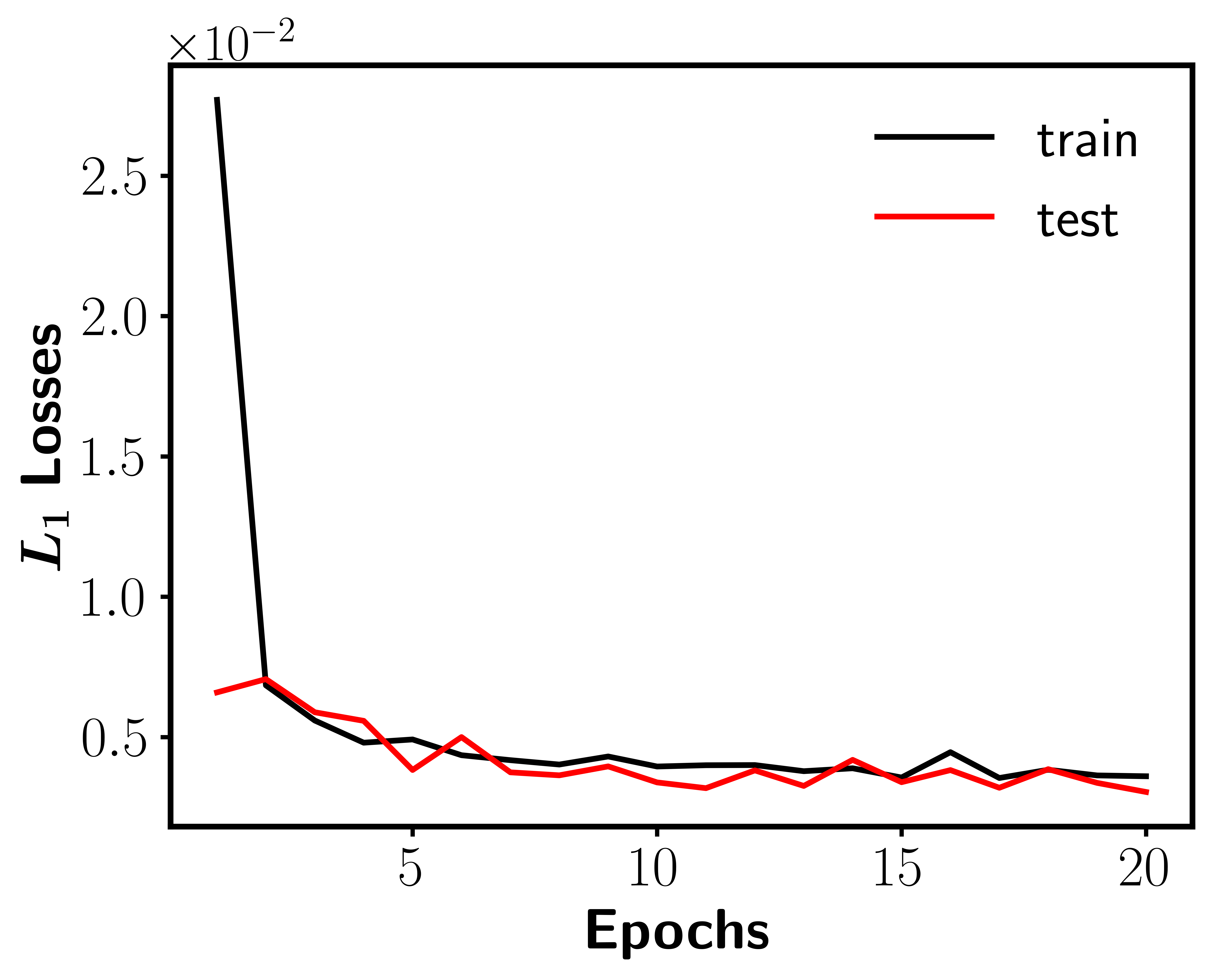}
		\end{minipage}%
		\hspace{0pt}
		\begin{minipage}{.4\textwidth}
			\centering
			\includegraphics[width=\linewidth]{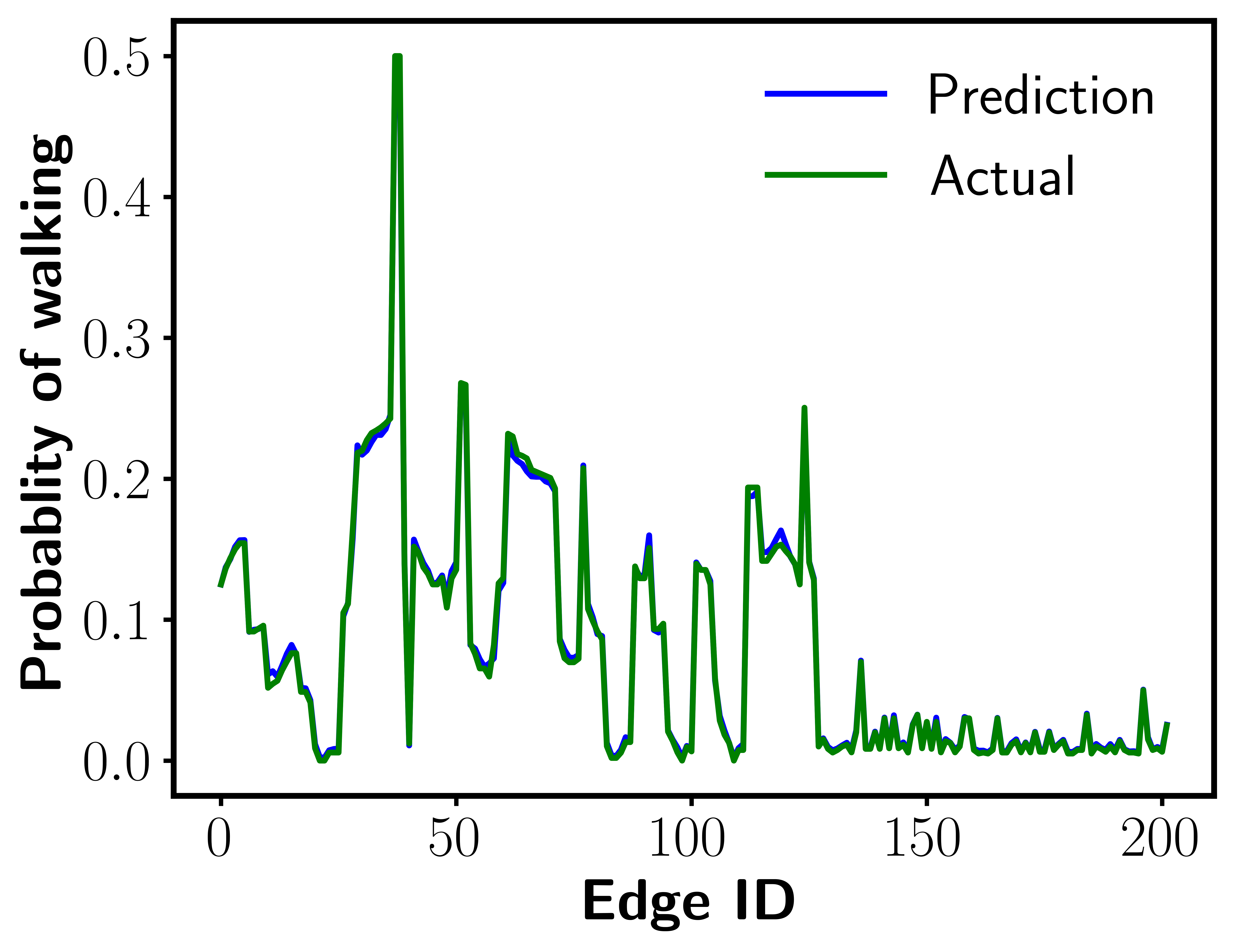}
		\end{minipage}
		\put(-6.5,2.0){\large Kushishang Mall}
	\end{picture}
	\vspace{3cm}
	\\
	\vspace{0pt}
	\begin{picture}(10,2.5)
		\begin{minipage}{.4\textwidth}
			\centering
			\includegraphics[width=\linewidth]{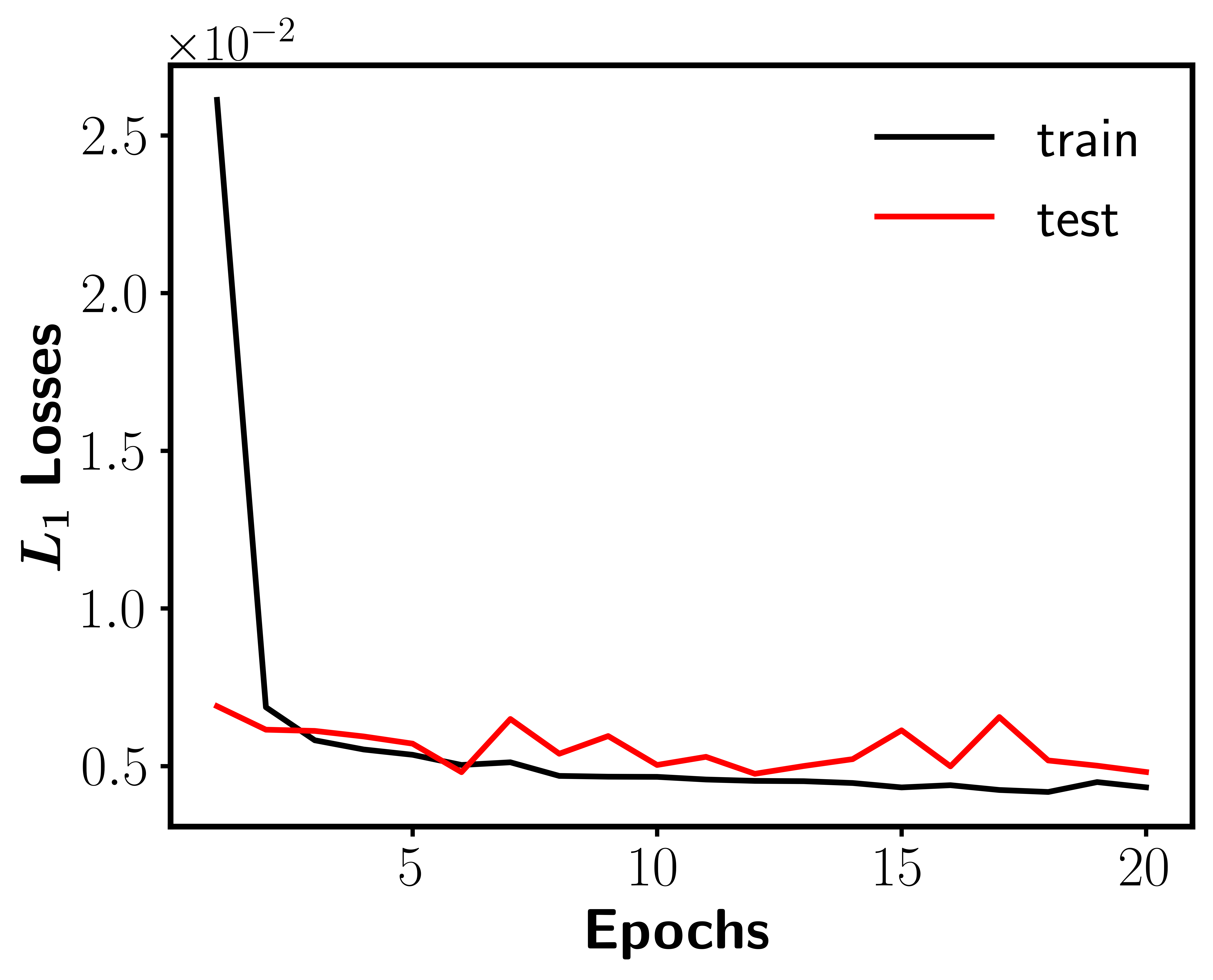}
		\end{minipage}%
		\hspace{0pt}
		\begin{minipage}{.4\textwidth}
			\centering
			\includegraphics[width=\linewidth]{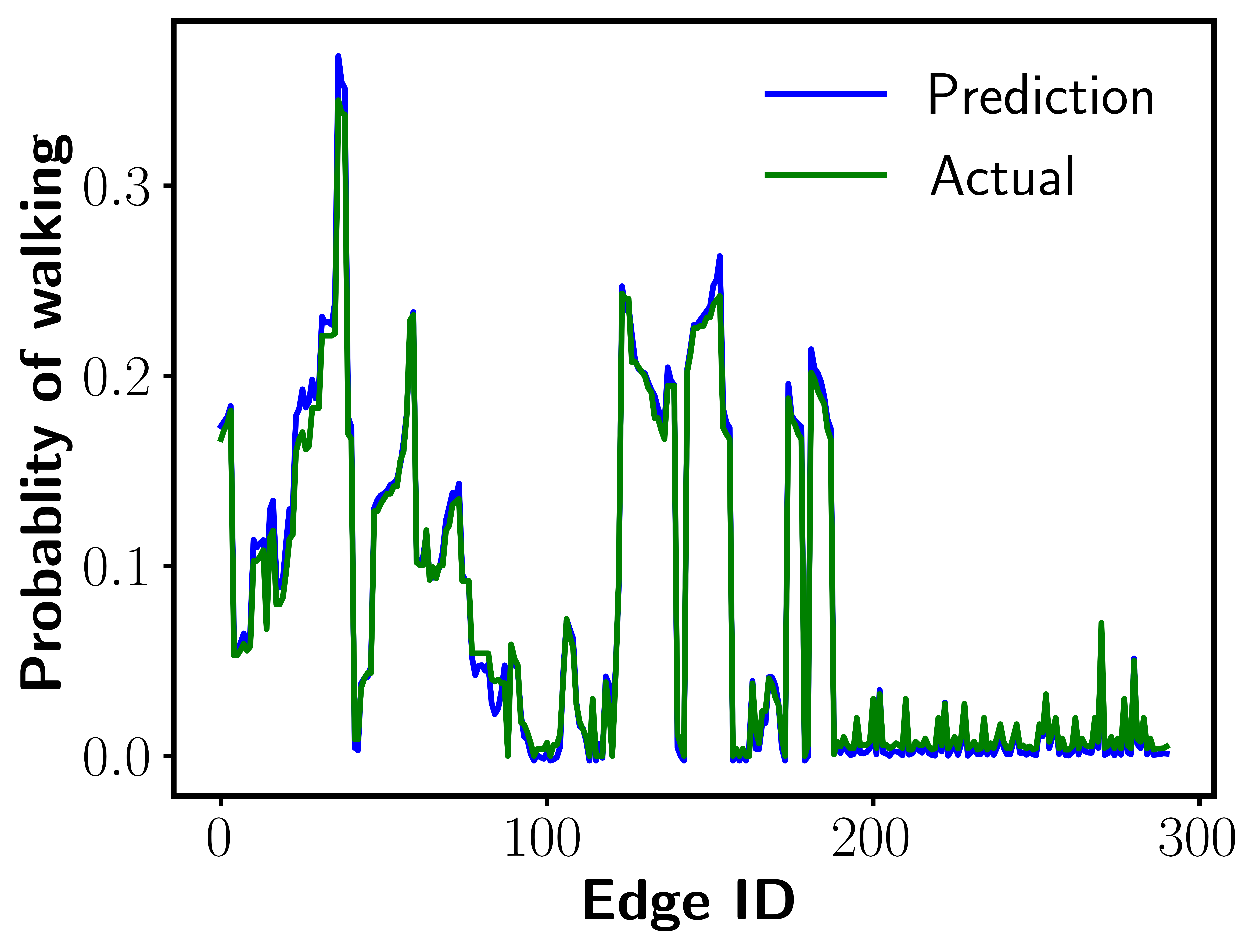}
		\end{minipage}
		\put(-6.5,2.0){\large Wanda Fengkedian Mall}
	\end{picture}
	\vspace{3cm}
	\\
	\vspace{0pt}
	\begin{picture}(10,2.5)
		\begin{minipage}{.4\textwidth}
			\centering
			\includegraphics[width=\linewidth]{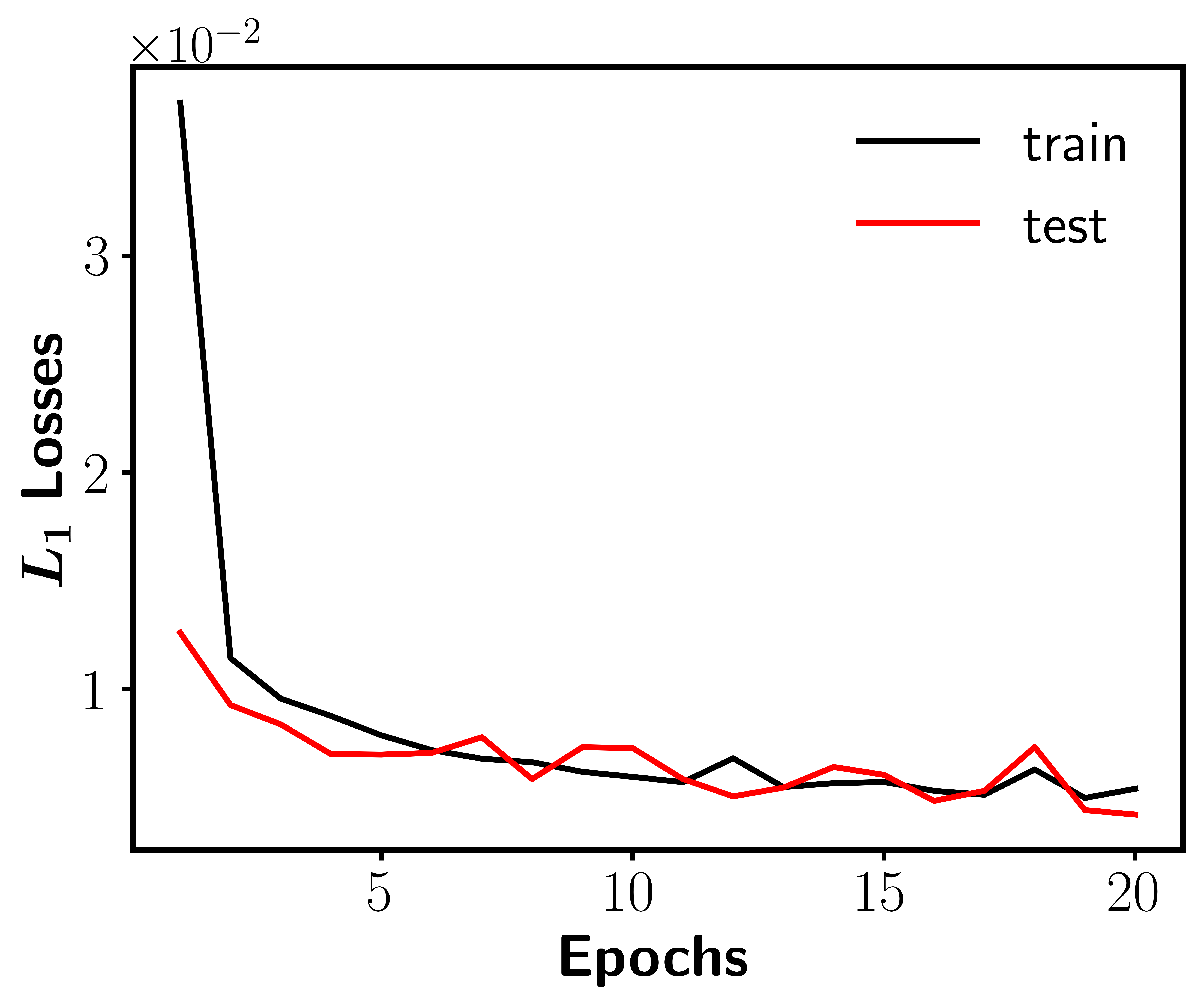}
		\end{minipage}%
		\hspace{0pt}
		\begin{minipage}{.4\textwidth}
			\centering
			\includegraphics[width=\linewidth]{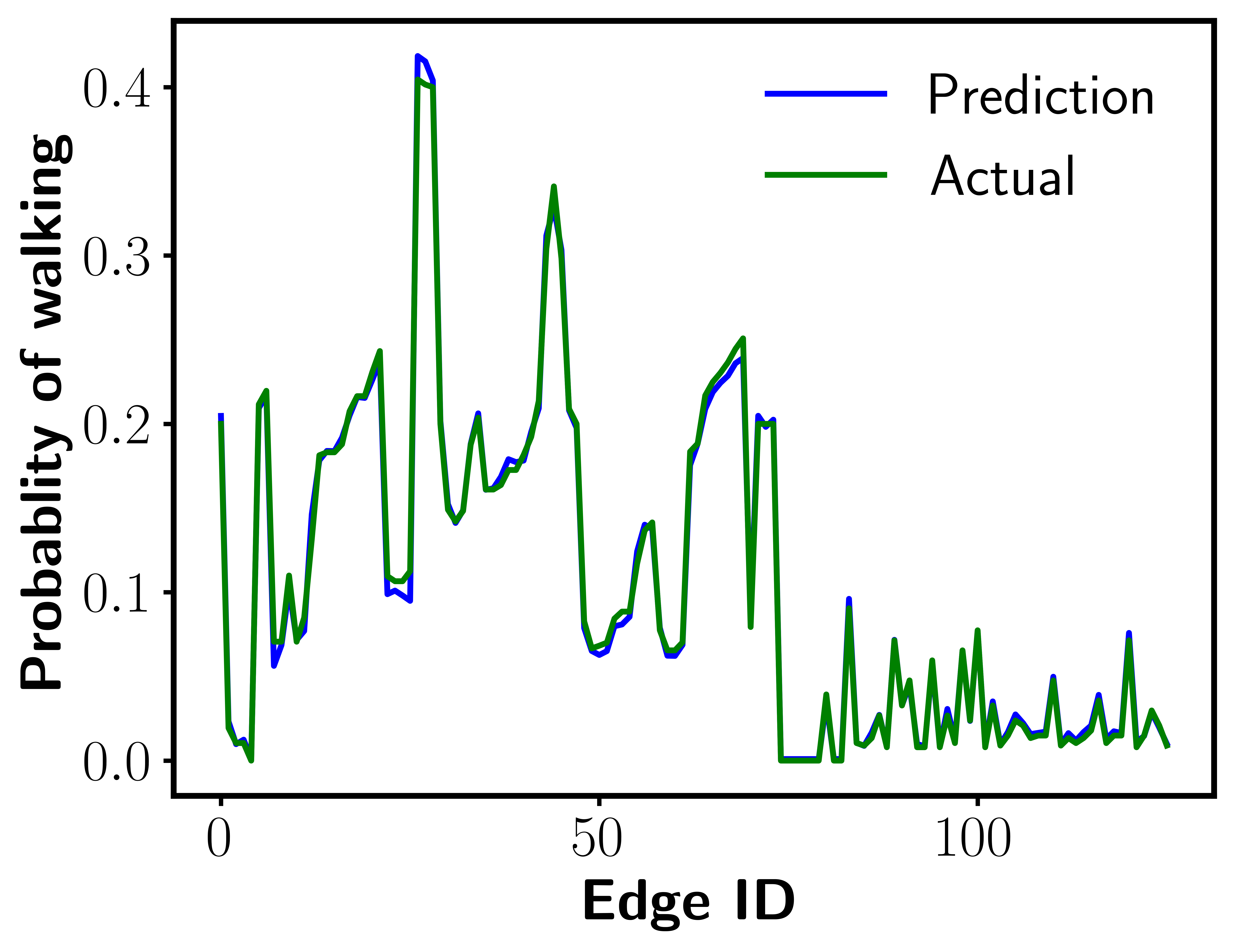}
		\end{minipage}
		\put(-6.5,2.0){\large Yintaibaihuo DHM Mall}
	\end{picture}
	\vspace{2.5cm}
	\caption{The training curves and the usage probabilities on the graph edges of three malls, namely, Kushishang, Wanda Fenkedian and Yintaibaihuo DHM: (left column) The training curve; (right column) The predicted and actual usage probabilities vs arbitrary edge ID's.}
	\label{fig:train_curve_pred_actual}
\end{figure*}

\begin{figure*}[htb!]
	\centering
	\vspace{0pt}
	\setlength{\unitlength}{0.08\textwidth}
	\begin{picture}(10,2.5)
		\begin{minipage}{.6\textwidth}
			\centering
			\includegraphics[width=\linewidth]{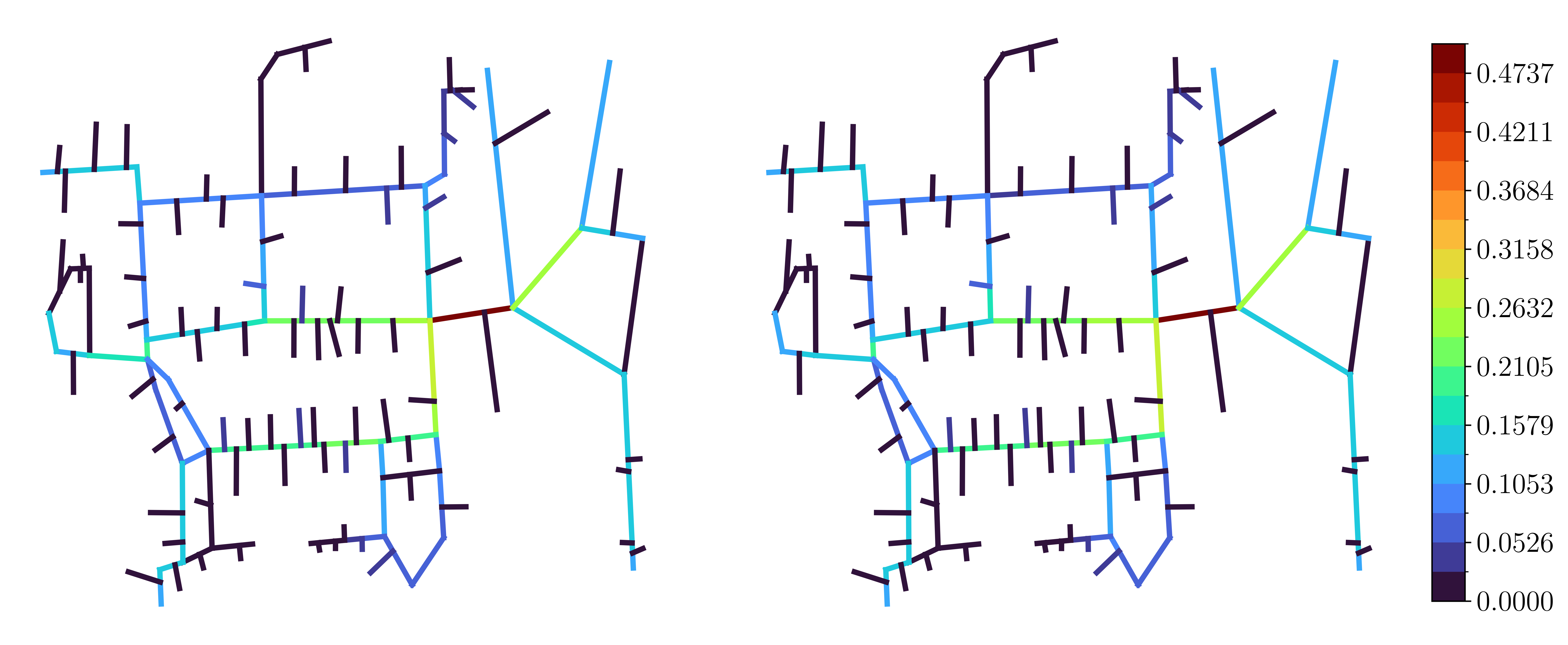}
		\end{minipage}%
		\hspace{0pt}
		\begin{minipage}{.3\textwidth}
			\centering
			\includegraphics[width=\linewidth]{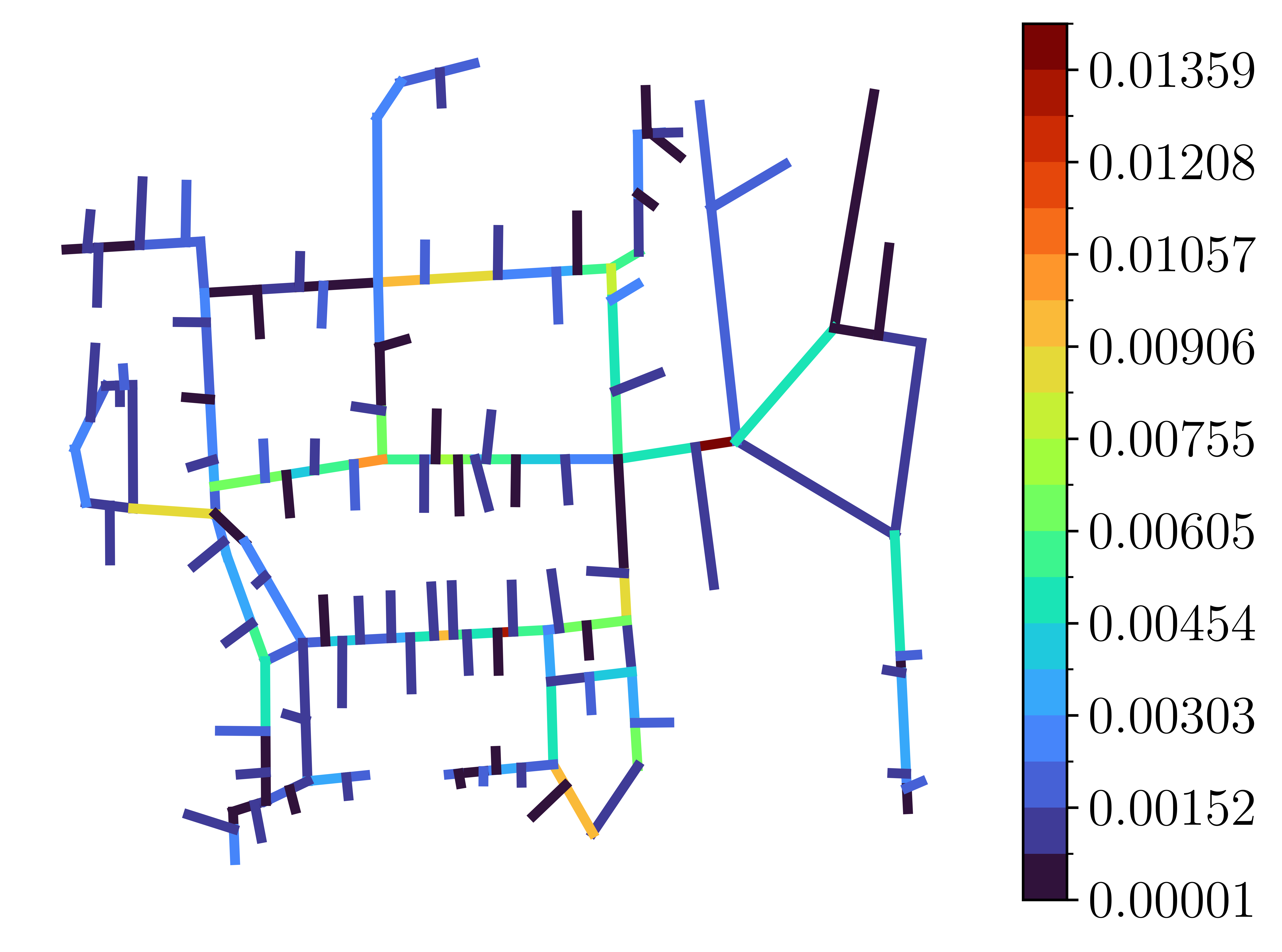}
		\end{minipage}
		\put(-6.5,2.0){\large Kushishang Mall}
	\end{picture}
	\vspace{3cm}
	\\
	\vspace{0pt}
	\begin{picture}(10,2.5)
		\begin{minipage}{.6\textwidth}
			\centering
			\includegraphics[width=\linewidth]{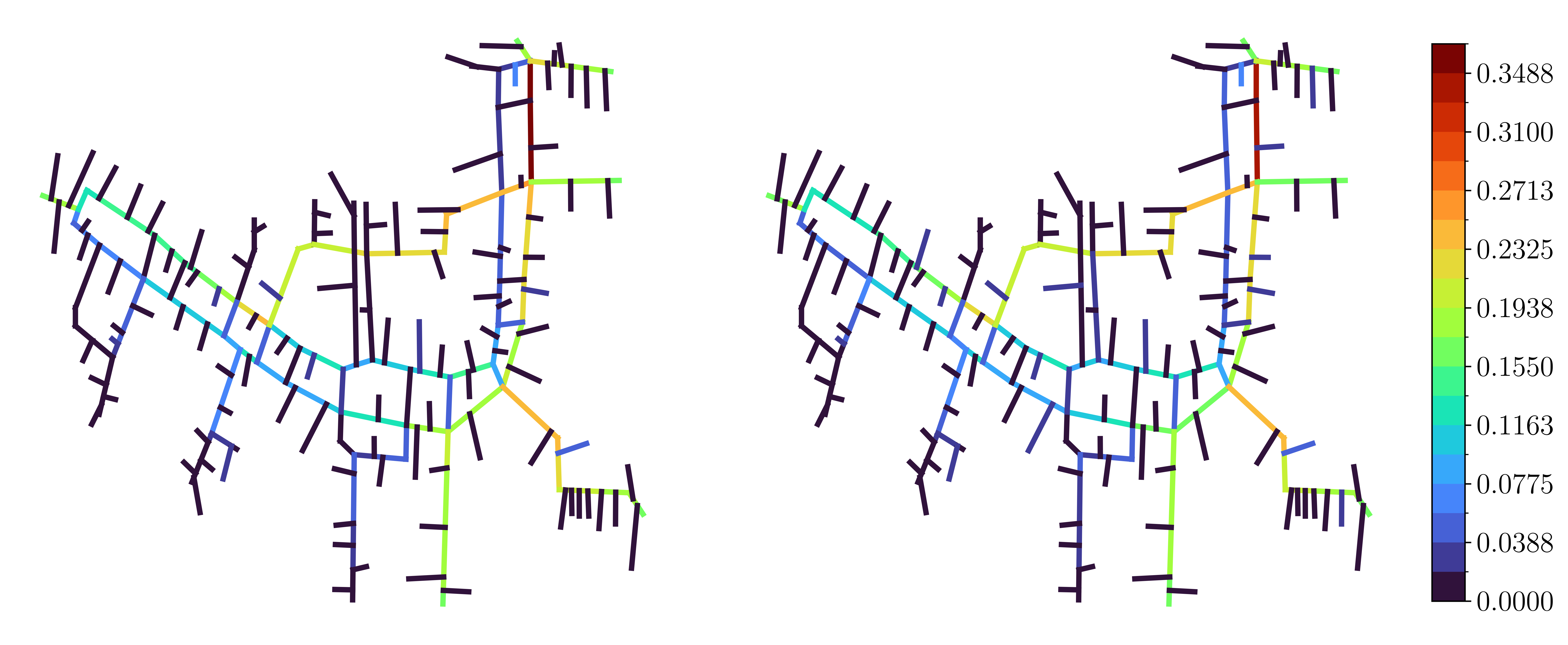}
		\end{minipage}%
		\hspace{0pt}
		\begin{minipage}{.3\textwidth}
			\centering
			\includegraphics[width=\linewidth]{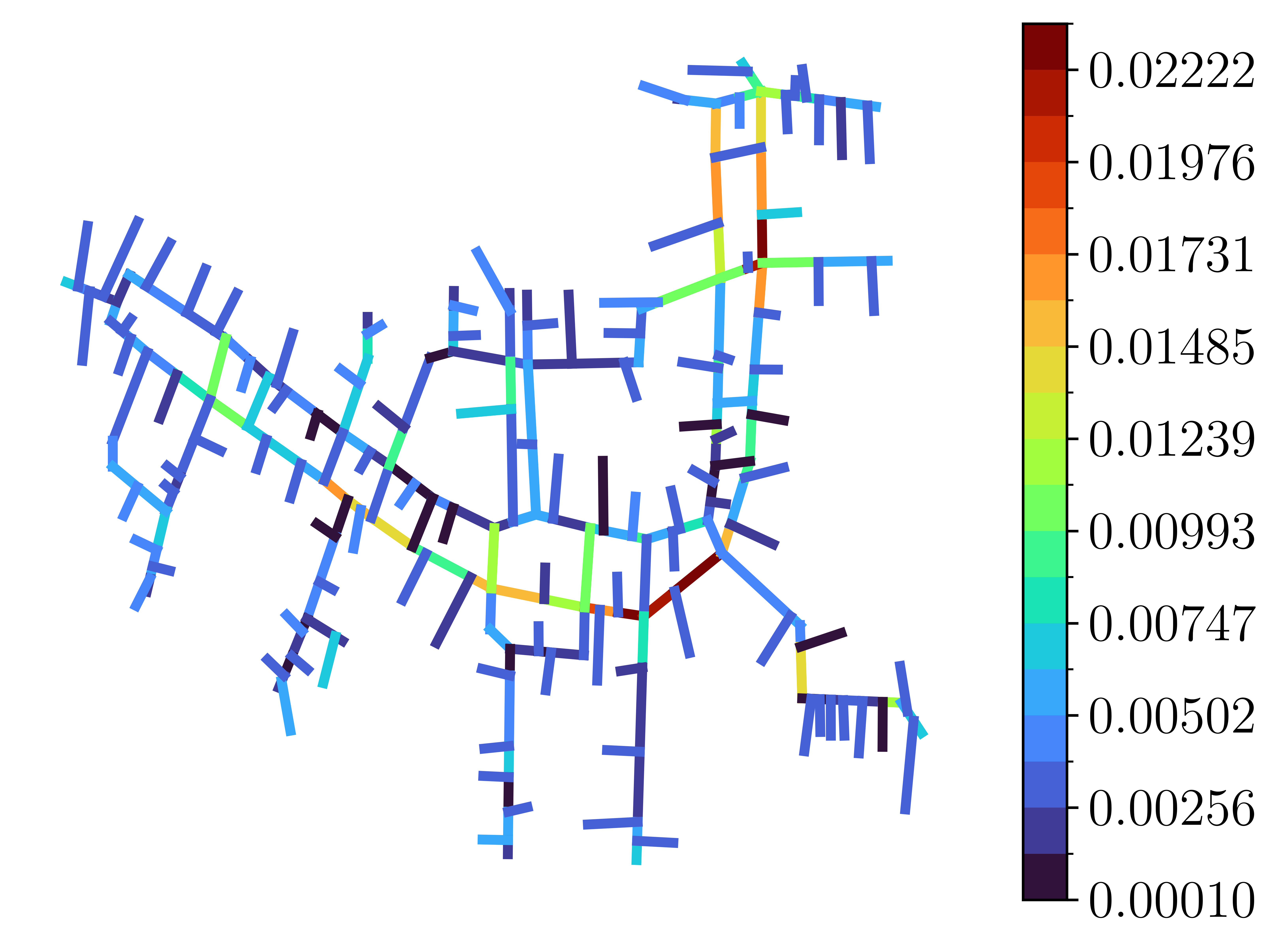}
		\end{minipage}
		\put(-6.5,2.0){\large Wanda Fengkedian Mall}
	\end{picture}
	\vspace{3cm}
	\\
	\vspace{0pt}
	\begin{picture}(10,2.5)
		\begin{minipage}{.6\textwidth}
			\centering
			\includegraphics[width=\linewidth]{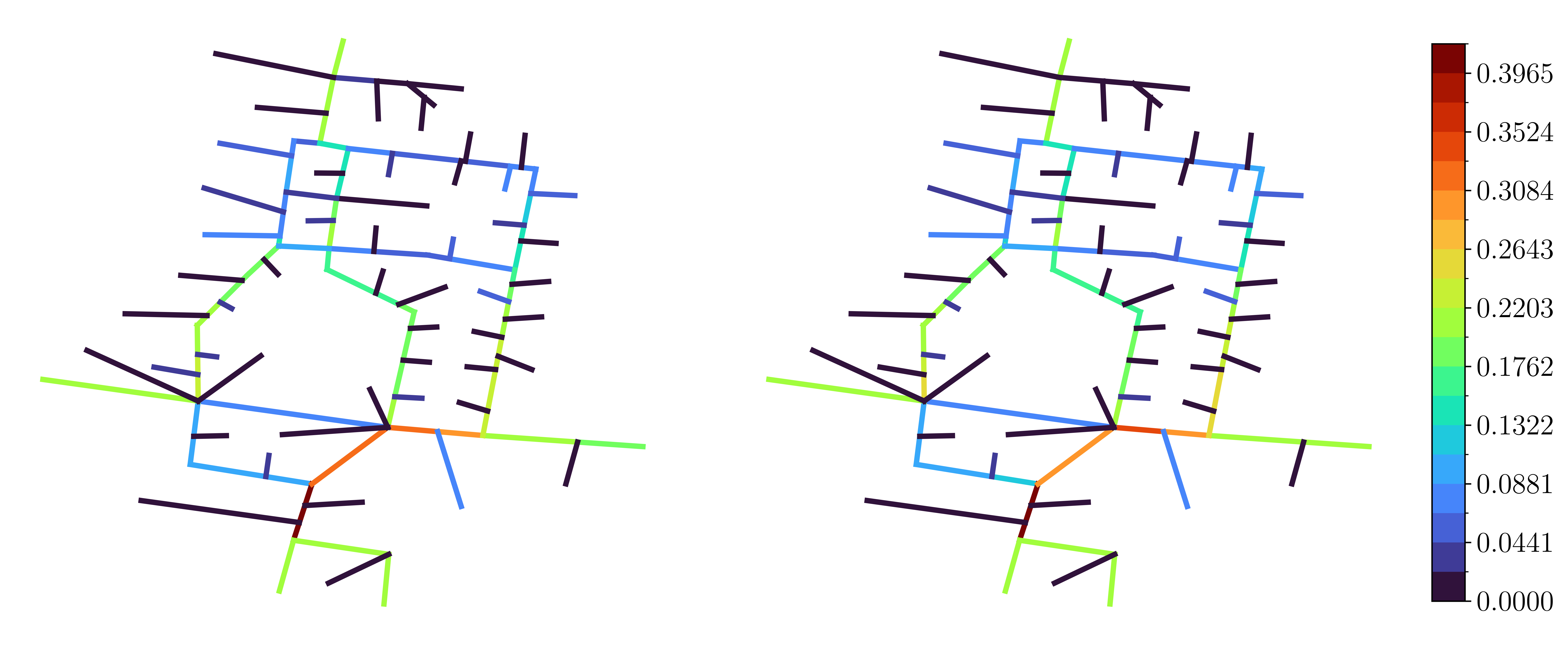}
		\end{minipage}%
		\hspace{0pt}
		\begin{minipage}{.3\textwidth}
			\centering
			\includegraphics[width=\linewidth]{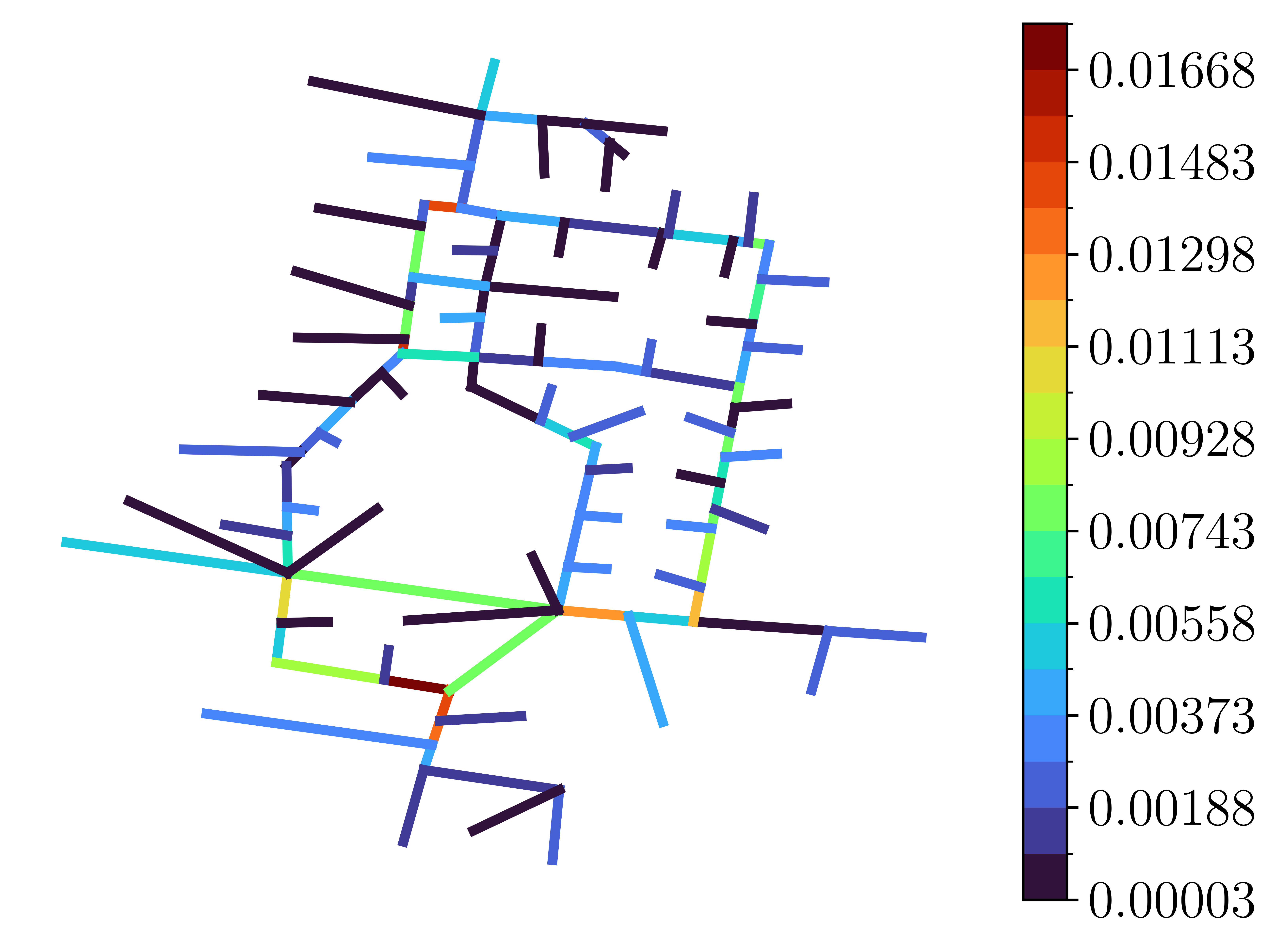}
		\end{minipage}
		\put(-6.5,2.0){\large Yintaibaihuo DHM Mall}
	\end{picture}
	\vspace{2.5cm}
	\caption{The usage probabilities on the graph edges of three malls, namely, Kushishang, Wanda Fenkedian and Yintaibaihuo DHM: (left column) The predicted; (centre column) The actual values given by the probability model; (right column) The mean absolute error between the left and centre columns. The color bar on the centre column also applies to the left column.}
	\label{fig:output_target_err}
\end{figure*}

\subsection{Main Model}
The main model, as an auto-encoder, essentially contains two parts: (i) an encoder formed by skip-connected seven heterogeneous GNN blocks, and (ii) a decoder named \emph{Predictor on Edges}, which is a block that converts the features on the nodes to those on the edges. This has also been designed to tackle the oversmoothing problem as described below in the subsection~\ref{subsec:oversmoothing}. 

The GNN problem that is solved by this main model can be described as below. In what follows we follow the notation introduced in the subsection~\ref{subsec:notation}. We have the heterogeneous graph $g(\boldsymbol{\mathcal{V}}, \boldsymbol{\mathcal{E}})$, where the set of nodes $\boldsymbol{\mathcal{V}}$ is a union of two types of nodes, i.e., $\boldsymbol{\mathcal{V}} = \boldsymbol{\mathcal{V}}^{(s)} \cup \boldsymbol{\mathcal{V}}^{(ns)}$, and the set of edges comprises two subsets as introduced before. Therefore, $\boldsymbol{\mathcal{E}} = \boldsymbol{\mathcal{E}}^{(s-ns)} \cup \boldsymbol{\mathcal{E}}^{(ns-ns)}$. These edges are featureless. Then, the model can be described as the function $\phi$ that outputs a tuple $y$ as shown in the Equation~(\ref{eq:main_model}).
\begin{equation}
	y = \phi(\boldsymbol{F}^{(s)}, \boldsymbol{F}^{(ns)};\ \boldsymbol{\mathcal{V}}^{(s)}, \boldsymbol{\mathcal{V}}^{(ns)}, \boldsymbol{\mathcal{E}}^{(s-ns)}, \boldsymbol{\mathcal{E}}^{(ns-ns)}),
	\label{eq:main_model}
\end{equation}
where $y \in \mathbb{R}^{\mathcal{E}}$, $\boldsymbol{F}^{(s)} \in \mathbb{R}^{\mathcal{V}^{(s)}\times 10}$ and $\boldsymbol{F}^{(ns)} \in \mathbb{R}^{\mathcal{V}^{(ns)}\times 10}$. Since the number of input features for shop and nonshop nodes are same, the Equation~(\ref{eq:main_model}) can be written as the mapping, $\phi: \mathbb{R}^{\mathcal{V}\times 10}\rightarrow \mathbb{R}^{\mathcal{E}}$. The function $\phi$ can be written as a composition of two functions as in
\begin{equation}
	\phi = \phi^{\textrm{DEC}} \circ \phi^{\textrm{ENC}},
	\label{eq:phiencdec}
\end{equation}
where the encoder $\phi^{\textrm{ENC}}: \mathbb{R}^{\mathcal{V}\times 10}\rightarrow \mathbb{R}^{\mathcal{V}\times n_h}$, the decoder $\phi^{\textrm{DEC}}: \mathbb{R}^{\mathcal{V}\times n_h}\rightarrow \mathbb{R}^{\mathcal{E}}$, and $n_h$ is the features of the hidden layers.

\subsection{Tackling Oversmoothing} \label{subsec:oversmoothing}
Since the graphs change between different shopping malls, we are in a need of a large number of learnable parameters in the encoder of the model to make accurate predictions. These large number of parameters can be expected to be more versatile in making predictions in a manner unaffected by the complexity of changing number of nodes and connections across the datasets. Since input vector is only 10 in length, we have to resort to increasing the number of layers in the model. However, increasing the GNN layers would lead to the well-known problem of graph-oversmoothing. In order to circumvent this from happening, we use few skip-connections that adds the features across previous layers and pass them to the subsequent layer. We found that a total of seven such GNN layers gives sufficient number of learnable parameters for this regression problem. Several layers as large as seven is quite high when compared to most GNN architectures, which use only 2 to 4 layers in general. However, as long as the oversmoothing problem is addressed by taking precautionary steps, a GNN model's accuracy has been observed to improve even when it is very deep with the number of layers as large as 20.~\cite{zhou2020towards}.

\subsection{Heterogeneous GNN Block}
This block comprises three main layers, namely, Graph convolution, batch normalization and a switch of type rectified linear unit (ReLU). 

The graph convolution layer can be one of the message-passing layers listed above. These layers take in features defined on each nodes of the graph at the input, perform message-passing convolutions with the feature vectors of the neighbouring nodes using some aggregation operations such as {\tt max} or {\tt mean}, and then they output a new set of features on each nodes. This way of message passing and subsequent aggregation help these layers to output features of each node by taking into account of their neighbours, a mechanism by which they outperform simple neural networks like multilayer perceptrons (MLP) which neglects the underlying graph connections and the permutability of the node id's the neighbours.

At the end of each GNN layers, the outputs are passed through a batch-normalization layer and the ReLU switch as shown in the diagram on the bottom right of Figure~\ref{fig:gnn_model}. However, as one can see from this block, the flow of information happens in two branches to accommodate two typed of edges in each of our graphs. Therefore, there are a pair of each of these layers in this block. 

However construction of this hetero-GNN block is simplified by building this model for a homogeneous graph and then by making use of the PyTorch Geometric's in-built tool, {\tt to\_heterogeneous()} that converts a homogeneous model to the heterogeneous model. 

Across the results that will be presented in next section, we have fixed the output size of each message passing heterogeneous-GNN block for both types of nodes as 16.

\subsection{Predictor on Edges} \label{poe}
Though some of the message-passing graph convolution layers have a version or options to treat edge-level features (for example, the length of each edge) that may be present in the graphs used in other studies, none of them could output features on the edge. However, the target data, i.e., the usage probabilities in our graphs live on edges. Therefore, we will need to use a separate block named \emph{predictor-on-edges} that outputs one each edge by taking into account of the node features on either end of it. These outputs on edges will be finally compared against the targets during training where the weights are tuned via back propagation.

The predictor-on-edges block is shown in top right of Figure~\ref{fig:gnn_model}. This block takes in the node-level features defined on each types of the nodes and a set of graph-level features. This is the decoder part of our GAE approach. 

In the traditional GAE approach, the decoder is a simple function that outputs a scalar as in the expression, $\mathrm{ReLU}(\mathbf{F}_i^{\mathrm{T}}\mathbf{F}_j)$, where $\mathbf{F}_i$ is the node feature of the $i^\mathrm{th}$ node represented as a column vector, and the superscript $\mathbf{T}$ stands for the transpose. Such a simple approach performs better in the link prediction tasks on large graphs. However, since this decoder lacks learnable parameters, it performs relatively poorer in our case which consists of datasets of graphs corresponding to different malls in each minibatch. Therefore, we modify the decoder to include a set of fully-connected layers to facilitate taking part in the training. This modification also help us to treat the graph-level features while outputting the final set of predictions.

Inside this block, for each edge of both types, namely, \emph{nonshop-nonshop} and \emph{shop-nonshop}, the product and mean of the node level features are computed and horizontally concatenated together with the graph level features. This concatenated quantity would serve as the edge-level features. We considered mean and product as these quantities are independent of commuting the operands, a feature required for the undirected graphs. 

We would like to highlight the reason why we consider the concatenated mean and product of the node-level features rather than the concatenated node-features themselves. The process of concatenation of node features directly breaks the symmetry of two nodes forming an edge, since one will be forced to choose one node's features as the first set of entries for the concatenation operation. Such symmetry is not broken when the results of sum and mean of the node features are concatenated.

Though the concatenated graph-level features do not change between different edges of the graph in same sample, or between the samples, they would change when compared between different shopping malls. This aspect is needed to make the model generalized for different graphs. However, this graph-level features can be ignored if the learning is based on the samples of a single mall. 

These edge features are organized further by concatenating them vertically such that the edge-type nonshop-to-nonshop is at the top. This ordering is made to conform with the order of the edges as they appear in the target data. 

The output from vertical-concatenation layer is passed through fully-connected (FC) layers as shown in Figure~\ref{fig:gnn_model} (top right). These FC layers help increasing the number of learnable parameters, as well as to gradually decrease the number of output features to one. Using the FC layers is a well-known strategy to increase the expressive power of a GNN model

\subsection{Training}
We show the predicted results after training from two types of datasets. In the first type, the learning is performed on the graphs of single mall. Since each mall is sampled 200 times, we used the splitting of 40 for test set, and 160 for training set.

Since the problem at our hand is of a regression type, we have mean-squared error (MSE) and $L1$-loss as candidates for the loss function. We used $L1$-loss, which is minimized by Adam optimizer~\cite{kingma2014adam} implemented in the deep-learning framework PyTorch. The number of learnable parameters of the model varied from 18,000 to 200,000 depending on the whether GNN layers (i.e., GraphSAGE, GAT or GIN) used. The training converged fairly in 20 epochs as can be seen from the learning curves shown in Figure~\ref{fig:train_curve_pred_actual} (left columns) for three malls whose names are mentioned on the figure. On the right column we show that on each of the edges the predicted and actual usage probabilities fairly match.

Figure~\ref{fig:output_target_err} shows the same results, but projected on the actual graphs of these three malls. The right-most column shows the mean-absolute error.

\section{Preliminary conclusion}
({\tt A full conclusion will appear in the next version of this article after including more results.}) The proposed GNN model has been found to work fairly well on the synthetic dataset prepared using a probability model. We also found that the prediction based on the combined dataset comprising the samples of different malls is also quite satisfactory, though these results will be included in our next iteration of this working paper.

({\tt To include Results from the combined dataset})

({\tt To include Results as a table comparing GraphSAGE, GAT and GIN})

({\tt To include more equations explaining the encoder and decoder.})

({\tt To include a section for Conclusion})

({\tt To include section without section number for code-availability})

\section*{Acknowledgment}
This research is supported by the National Research Foundation, Prime Minister's Office, Singapore under its Cities of Tomorrow R\&D Programme (COT-H1-2020-2). Any opinions, findings and conclusions or recommendations expressed in this material are those of the author(s) and do not reflect the views of National Research Foundation, Singapore and Ministry of National Development, Singapore.



\bibliographystyle{elsarticle-num} 
\bibliography{gnn_pedestrian}

%
%
%
%

\end{document}